\documentclass[a4paper]{easychair}

\usepackage{doc}

\title{Efficient and Reliable Hitting-Set Computations\\ for the Implicit Hitting Set Approach
}

\author{
	Hannes Ihalainen\inst{1} 
	\and 
	Dieter Vandesande\inst{2,3}
	\and 
	André Schidler\inst{4}
	\and\\  Jeremias Berg\inst{1}
	\and Bart Bogaerts\inst{3,2}
	\and Matti Järvisalo\inst{1} 
}

\institute{
  University of Helsinki, Dept.\ Computer Science, Helsinki, Finland
\and
Vrije Universiteit Brussel, Dept.\ Computer Science, Brussels, Belgium
\and
KU Leuven, Dept.\ Computer Science, Leuven, Belgium
\and
University of Freiburg, Freiburg, Germany
 }

\authorrunning{Ihalainen et al}

\titlerunning{Efficient and Reliable Hitting-Set Computations for the Implicit Hitting Set Approach}

\usepackage{xspace}
\usepackage{amssymb,amsmath}
\usepackage{todonotes}
\usepackage{booktabs}

\usepackage{amsthm}
\newtheorem{theorem}{Theorem}
\newtheorem{proposition}[theorem]{Proposition}
\newtheorem{example}[theorem]{Example}

\usepackage{graphicx}
\usepackage[linesnumbered,noend]{algorithm2e}
\usepackage[noend]{algorithmic}
\usepackage{float}

\usepackage{xcolor, soul}% <<<
\sethlcolor{yellow}% <<<

\newcommand\litell{\ell}
\newcommand\varx{x\xspace}
\newcommand\olnot[1]{\overline{#1}\xspace}
\newcommand{\coeffa}{a\xspace}
\newcommand\weight{w\xspace}
\newcommand\dega{A\xspace}
\newcommand{\constrC}{C\xspace}
\newcommand\constrc{\constrC}
\newcommand{\formF}{F\xspace}
\newcommand{\formf}{\formF}
\newcommand{\objective}{O}
\newcommand\bigand{\bigwedge}
\newcommand{\assmnt}{\alpha}
\newcommand\assmnttwo{\beta}
\newcommand{\sumnodisplay}{{\textstyle \sum}}

\newcommand\lowerbound{lb}
\newcommand\upperbound{ub}

\newcommand{\ubModel}{\assmnt_{\mathit{best}}}
\newcommand{\true}{\textsc{true}\xspace}
\newcommand{\coreconstr}{\constrC}

\newcommand{\csol}{\gamma}

\newcommand\subst{\omega}
\newcommand{\accumCores}{\mathcal{K}}
\newcommand{\newCores}{\mathcal{N}}

\SetKwProg{algo}{}{}{}
\SetKwFunction{ihs}{PBO-IHS}
\newcommand\ihsInTitle\ihs
\SetKwFunction{ihssym}{PBO-IHS-w-SCL}
\SetKwFunction{compsyms}{Compute-Symmetries}
\SetKwFunction{filterSymmetries}{Filter-Symmetries}
\SetKwFunction{addSumVariable}{Add-Count-Variable}
\SetKwFunction{computeOrbit}{Compute-Orbit}
\SetKwFunction{addSumVariableI}{Add-Count-Variables}
\SetKwFunction{computeOrbitI}{Find-Interchangeable-Lits}
\SetKwFunction{selectSubsets}{Select-Candidate-Partitions}
\SetKwFunction{extendInCore}{Extend-In-Core}
\SetKwFunction{extendOutCore}{Extend-Outside-Core}
\SetKwFunction{checkOrbit}{Check-Interch}
\SetKwFunction{symlearn}{SCL}
\SetKwFunction{pb}{PB-Solve}
\SetKwFunction{oracle}{Extract-Cores}
\SetKwFunction{mchs}{Min-Sol}
\SetKwFunction{solvehs}{Solve-HS}
\newcommand\solvehsInTitle\solvehs
\SetKwFunction{usesls}{use-sls}
\SetKwFunction{optimalsol}{optimal-sol}
\SetKwFunction{optimizersol}{Optimize-UB}
\SetKwFunction{optimizermin}{Optimize}
\newcommand\optimizerminInTitle\optimizermin
\SetKwFunction{sls}{Core-SLS}

\SetKwFunction{explicitlearn}{Explicit-SCL}
\SetKwFunction{compactlearn}{Compact-SCL}
\SetKwFunction{trycompactlearn}{Try-Compact-SCL}
\SetKwFunction{heuristicSkip}{Check-If-Good}
\SetKwFunction{seeding}{seed}

\newcommand\veripb{\textsc{VeriPB}\xspace}

\renewcommand\optimizerminInTitle{Optimize\xspace}
\renewcommand\ihsInTitle{PBO-IHS\xspace}
\renewcommand\solvehsInTitle{Solve-HS\xspace}
\usepackage{etoolbox}

\ExplSyntaxOn

\newcommand\setcitation[2]{%
	\csdef{mycommoncitation\text_uppercase:n{#1}}{#2}%
	\csappto{bbAllCommonCitations}{\cite{#2}\ }%%%for testing purpuses only	
}
\newcommand\getcitation[1]{%
	\csuse{mycommoncitation\text_uppercase:n{#1}}}

\newcommand\refto[1]{%
	\ifcsname  mycommoncitation\text_uppercase:n{#1}\endcsname%
	\getcitation{#1}%
	\else%
	#1%
	\fi%
}

\ExplSyntaxOff

\newcommand\mycite[1]{\cite{\refto{#1}}}

\setcitation{IDP}{DBBJD18PredicatelogicmodelinglanguageIDPsystem}
\setcitation{fodot}{DT08logicnonmonotoneinductivedefinitions}
\setcitation{CPsupport}{DBDD13ModelExpansionPresenceFunctionSymbolsUsing}
\setcitation{minisatid}{DBDD13ModelExpansionPresenceFunctionSymbolsUsing}
\setcitation{FunctionDetection}{DB13Detectionexploitationfunctionaldependenciesmodelgeneration}
\setcitation{lazyGrounding}{DDBS15LazyModelExpansionInterleavingGroundingSearch} 
\setcitation{Bootstrapping}{BJDJBD16BootstrappingInferenceIDPKnowledgeBaseSystem}
\setcitation{BootstrappingIDP}{BJDJBD16BootstrappingInferenceIDPKnowledgeBaseSystem}
\setcitation{FOSymmetry}{DBBD16localdomainsymmetrymodelexpansion}
\setcitation{FOLASP}{VDV21FOLASPFOInputLanguageAnswerSet}
\setcitation{inputster}{JJJ13CompilingInputFOinductivedefinitionsinto}
\setcitation{LearningPaper}{BBBDDJLR15Predicatelogicmodelinglanguagemodelingsolving}

\setcitation{foid}{DT08logicnonmonotoneinductivedefinitions}
\setcitation{cplogic}{VDB10FOIDextensionDLrules}
\setcitation{clog}{BVDV14FOCKnowledgeRepresentationLanguageCausality} %TODO replace by journal publication if one is published
\setcitation{foc}{BVDV14FOCKnowledgeRepresentationLanguageCausality}
\setcitation{inferenceClog}{BVDV14InferenceFOCModellingLanguage}
\setcitation{inferenceFOC}{BVDV14InferenceFOCModellingLanguage}
\setcitation{examplesClog}{BVDV14FOCRelatedModellingParadigms} %TODO replace by
\setcitation{satid}{MWDB08SATIDSatisfiabilityPropositionalLogicExtendedInductive}

\setcitation{fodot2asp}{DLTV19informalsemanticsAnswerSetProgrammingTarskian} %TODO replace by journal publication if one is publishedr
\setcitation{Tarskian}{DLTV19informalsemanticsAnswerSetProgrammingTarskian} %Same as the one above 
\setcitation{TarskianSemanticsASP}{DLTV19informalsemanticsAnswerSetProgrammingTarskian} %Same as the one above 
\setcitation{KBS}{DV08BuildingKnowledgeBaseSystemIntegrationLogic}
\setcitation{KBS-invitedtalk}{D16FOKnowledgeBaseSystemproject}
\setcitation{KBPE}{DWD11PrototypeKnowledge-BasedProgrammingEnvironment}

\setcitation{justifications}{DBS15FormalTheoryJustifications} 
\setcitation{justificationTheory}{DBS15FormalTheoryJustifications} 
\setcitation{AFT}{DMT00ApproximationsStableOperatorsWell-FoundedFixpointsApplications}

\setcitation{SAT}{BHvW21HandbookSatisfiability-SecondEdition}
\setcitation{HandbookOfSAT}{BHvW21HandbookSatisfiability-SecondEdition}
\setcitation{SMS}{KS24SATModuloSymmetriesGraphGenerationEnumeration}

\setcitation{totalizer}{BB03EfficientCNFEncodingBooleanCardinalityConstraints}

\setcitation{satsuma}{ABR24satsumaStructure-BasedSymmetryBreakingSAT}
\setcitation{BreakID}{DBBD16ImprovedStaticSymmetryBreakingSAT}
\setcitation{symchaff}{S09SymChaffexploitingsymmetrystructure-awaresatisfiabilitysolver}

\setcitation{saucy}{KSM10SymmetrySatisfiabilityUpdate}
\setcitation{dejavu}{AS21EngineeringFastProbabilisticIsomorphismTest}
\setcitation{nauty}{MP14PracticalgraphisomorphismII}
\setcitation{bliss}{JK07EngineeringEfficientCanonicalLabelingToolLarge}

\setcitation{DPLLT}{GHNOT04DPLLTFastDecisionProcedures}
\setcitation{SMT}{BSST21SatisfiabilityModuloTheories}

\setcitation{CP}{RvW06HandbookConstraintProgramming}
\setcitation{csp2asp}{DW11Translation-BasedConstraintAnswerSetSolving}
\setcitation{EZCSP}{B11Industrial-SizeSchedulingASPCP}
\setcitation{Inca}{DW12AnswerSetSolvingLazyNogoodGeneration}
\setcitation{Zinc}{MNRSGW08DesignZincModellingLanguage}
\setcitation{MiniZinc}{NSBBDT07MiniZincTowardsStandardCPModellingLanguage}

\setcitation{ASPComp2}{DVBGT09SecondAnswerSetProgrammingCompetition}
\setcitation{ASPComp3}{CIR14thirdopenanswersetprogrammingcompetition}
\setcitation{ASPComp4}{ACCDDIKK13FourthAnswerSetProgrammingCompetitionPreliminary}
\setcitation{ASPComp5}{CGMR16DesignresultsFifthAnswerSetProgramming}
\setcitation{ASPComp6}{GMR17SixthAnswerSetProgrammingCompetition}
\setcitation{ASPComp7}{GMR20SeventhAnswerSetProgrammingCompetitionDesign}
\setcitation{AspInPractice}{GKKS12AnswerSetSolvingPractice}
\setcitation{clasp}{GKS12Conflict-drivenanswersetsolvingFromtheory}
\setcitation{oclingo}{GGKOSS12StreamReasoningAnswerSetProgrammingPreliminary}
\setcitation{clingo}{GKKOSW16TheorySolvingMadeEasyClingo5}
\setcitation{gringo}{GST07GrinGoNewGrounderAnswerSetProgramming}
\setcitation{cmodels}{GLM04SAT-BasedAnswerSetProgramming}
\setcitation{DLV}{LPFEGPS06DLVsystemknowledgerepresentationreasoning}
\setcitation{GroundingBottleneck}{SPL14TwoApplicationsASP-PrologSystemDecomposablePrograms}
\setcitation{lazygroundingASP}{BW18ExploitingJustificationsLazyGroundingAnswerSet} 
\setcitation{justificationsAlpha}{BW18ExploitingJustificationsLazyGroundingAnswerSet}
\setcitation{ASP}{MT99StableModelsAlternativeLogicProgrammingParadigm}
\setcitation{OLL}{AKMS12Unsatisfiability-basedoptimizationclasp} %{MT99StableModelsAlternativeLogicProgrammingParadigm,N99LogicProgramsStableModelSemanticsConstraint,L99AnswerSetPlanning} 

\setcitation{Hilog}{CKW93HILOGFoundationHigher-OrderLogicProgramming}

\setcitation{MaxSAT}{LM21MaxSATHardSoftConstraints}

\setcitation{KR}{vLP08HandbookKnowledgeRepresentation}

\setcitation{lazyclausegeneration}{OSC09Propagationlazyclausegeneration}
\setcitation{FP}{H89ConceptionEvolutionApplicationFunctionalProgrammingLanguages}
\setcitation{GroundingWithBounds}{WMD10GroundingFOFOIDBounds}
\setcitation{GroundWithBounds}{WMD10GroundingFOFOIDBounds}
\setcitation{LTC}{BJBDVD14SimulatingDynamicSystemsUsingLinearTime}
\setcitation{SPSAT}{DBDDM12SymmetryPropagationImprovedDynamicSymmetryBreaking}
\setcitation{LCG}{S10LazyClauseGenerationCombiningPowerSAT}

\setcitation{GroundedFixpoints}{BVD15Groundedfixpointstheirapplicationsknowledgerepresentation}
\setcitation{PartialGroundedFixpoints}{BVD15PartialGroundedFixpoints}
\setcitation{LogicBlox}{GAK12LogicBloxPlatformLanguageTutorial}
\setcitation{proB}{LB08ProBautomatedanalysistoolsetBmethod}
\setcitation{NaturalInductions}{DV14Well-FoundedSemanticsPrincipleInductiveDefinitionRevisited} %TODO replace by journal publication if one is published
\setcitation{LP}{vK76SemanticsPredicateLogicProgrammingLanguage}

\setcitation{AF}{D95AcceptabilityArgumentsFundamentalRoleNonmonotonicReasoning}
\setcitation{ADF}{BW10AbstractDialecticalFrameworks}
\setcitation{ADFRevisited}{BSEWW13AbstractDialecticalFrameworksRevisited}
\setcitation{DefaultLogic}{R80LogicDefaultReasoning}
\setcitation{DL}{R80LogicDefaultReasoning}
\setcitation{AEL}{M85SemanticalConsiderationsNonmonotonicLogic}
\setcitation{minisat}{ES03ExtensibleSAT-solver}
\setcitation{completion}{C77NegationFailure}
\setcitation{ClarkCompletion}{C77NegationFailure}
\setcitation{wasp}{ADFLR13WASPNativeASPSolverBasedConstraint}
\setcitation{CEGAR}{CGJLV03Counterexample-guidedabstractionrefinementsymbolicmodelchecking}
\setcitation{CuttingPlane}{DFJ54SolutionLarge-ScaleTraveling-SalesmanProblem}
\setcitation{CuttingPlanes}{DFJ54SolutionLarge-ScaleTraveling-SalesmanProblem}
\setcitation{kodkod}{TJ07KodkodRelationalModelFinder}
\setcitation{CDCL}{SS99GRASPSearchAlgorithmPropositionalSatisfiability}
\setcitation{1UIP}{ZMMM01EfficientConflictDrivenLearningBooleanSatisfiability}
\setcitation{relevance}{JBDJD16RelevanceSATID}
\setcitation{relevance-implementation}{JDBJD16ImplementingRelevanceTrackerModule}
\setcitation{WFS}{VRS91Well-FoundedSemanticsGeneralLogicPrograms}
\setcitation{UnfoundedSet}{VRS91Well-FoundedSemanticsGeneralLogicPrograms}
\setcitation{UFS}{VRS91Well-FoundedSemanticsGeneralLogicPrograms}
\setcitation{stablesemantics}{GL88StableModelSemanticsLogicProgramming}
\setcitation{shatter}{ASM06EfficientSymmetryBreakingBooleanSatisfiability}
\setcitation{sbass}{DTW11Symmetry-breakinganswersetsolving}
\setcitation{lparsemanual}{url:lparsemanual}
\setcitation{AIC}{FGZ04Activeintegrityconstraints}
\setcitation{templates}{DvJD15Semanticstemplatescompositionalframeworkbuildinglogics}
\setcitation{templates2}{DvBJD16CompositionalTypedHigher-OrderLogicDefinitions}
\setcitation{sat-to-sat}{JTT16SAT-to-SATDeclarativeExtensionSATSolversNew}
\setcitation{sat-to-sat-qbf}{BJT16SolvingQBFInstancesNestedSATSolvers}
\setcitation{sat-to-sat-SO}{BJT16DeclarativeSolverDevelopmentCaseStudies}
\setcitation{XSB}{SW12XSBExtendingPrologTabledLogicProgramming}
\setcitation{KCmap}{DM02KnowledgeCompilationMap}
\setcitation{KnowledgeCompilationMap}{DM02KnowledgeCompilationMap}
\setcitation{TLA}{L02SpecifyingSystemsTLALanguageToolsHardware}
\setcitation{EventB}{A10ModelingEvent-B-SystemSoftwareEngineering}
\setcitation{MX}{MT05FrameworkRepresentingSolvingNPSearchProblems}
\setcitation{MIP}{S96Linearintegerprogramming-theorypractice}
\setcitation{PerfectModel}{P88DeclarativeSemanticsDeductiveDatabasesLogicPrograms}
\setcitation{SafeInductions}{BVD17SafeInductionsAlgebraicStudy}
\setcitation{alpha}{W17BlendingLazy-GroundingCDNLSearchAnswer-SetSolving}
\setcitation{omiga}{DEFWW12OMiGAOpenMindedGroundingOn-The-FlyAnswer}
\setcitation{gasp}{PDPR09GASPAnswerSetProgrammingLazyGrounding}
\setcitation{asperix}{LN09FirstVersionNewASPSolverASPeRiX}
\setcitation{CTL}{CE81DesignSynthesisSynchronizationSkeletonsUsingBranching-Time}
\setcitation{AFT-AIC}{BC18Fixpointsemanticsactiveintegrityconstraints}
\setcitation{UltimateApproximator}{DMT04Ultimateapproximationapplicationnonmonotonicknowledgerepresentation}
\setcitation{KripkeKleene}{F85Kripke-KleeneSemanticsLogicPrograms}
\setcitation{AFT-HO}{CRS18ApproximationFixpointTheoryWell-FoundedSemanticsHigher-Order} 
\setcitation{HereThere}{H30formalenRegelnintuitionistischenLogik}
\setcitation{dAEL}{VCBD16DistributedAutoepistemicLogicApplicationAccessControl}
\setcitation{SDD}{D11SDDNewCanonicalRepresentationPropositionalKnowledge}
\setcitation{HEX}{EIST06dlvhexProverSemantic-WebReasoningunderAnswer-Set}
\setcitation{wADF}{BSWW18WeightedAbstractDialecticalFrameworks}
\setcitation{wADFfix}{BPSWW18WeightedAbstractDialecticalFrameworksExtendedRevised}
\setcitation{TransitionSystems}{NOT06SolvingSATSATModuloTheoriesFrom}
\setcitation{galliwasp}{MG12GalliwaspGoal-DirectedAnswerSetSolver}
\setcitation{clingcon}{BKOS17Clingconnextgeneration}
\setcitation{lp2sat}{JN11CompactTranslationsNon-disjunctiveAnswerSetPrograms}
\setcitation{lp2mip}{LJN12AnswerSetProgrammingMixedIntegerProgramming}
\setcitation{lp2diff}{JNS09ComputingStableModelsReductionsDifferenceLogic}
\setcitation{lp2acyc}{GJR14AnswerSetProgrammingSATmoduloAcyclicity}
\setcitation{PB}{RM09Pseudo-BooleanCardinalityConstraints}
\setcitation{CuttingPlanes}{CCT87complexitycutting-planeproofs}
\setcitation{RoundingSAT}{EN18DivideConquerTowardsFasterPseudo-BooleanSolving}
\setcitation{PRS}{DG02InferenceMethodsPseudo-BooleanSatisfiabilitySolver}
\setcitation{sat4j}{LP10Sat4jlibraryrelease22}
\setcitation{mingo}{LJN12AnswerSetProgrammingMixedIntegerProgramming}
\setcitation{pbmodels}{LT05Pbmodels-SoftwareComputeStableModels}
\setcitation{HEF-LP}{GLL07Head-Elementary-Set-FreeLogicPrograms}
\setcitation{HCF-LP}{BD94PropositionalSemanticsDisjunctiveLogicPrograms}
\setcitation{aspcore2}{CFGIKKLM20ASP-Core-2InputLanguageFormat}
\setcitation{SemanticWeb}{AGvH12SemanticWebPrimer3rdEdition}
\setcitation{QBF}{KB21TheoryQuantifiedBooleanFormulas}
\setcitation{SMUS}{IPLM15SmallestMUSExtractionMinimalHittingSet}
\setcitation{CDCLsym}{MBCK18CDCLSymIntroducingEffectiveSymmetryBreakingSAT}
\setcitation{SEL}{DBB17SymmetricExplanationLearningEffectiveDynamicSymmetry}
\setcitation{Coq}{BC04InteractiveTheoremProvingProgramDevelopment-}
\setcitation{Isabelle}{NPW02IsabelleHOL-ProofAssistantHigher-OrderLogic}
\setcitation{HOL}{SN08BriefOverviewHOL4}
\setcitation{lean}{dU21Lean4TheoremProverProgrammingLanguage}
\setcitation{CakeML}{TMKFON19verifiedCakeMLcompilerbackend}
\setcitation{FRAT}{BCH22FlexibleProofFormatSATSolver-ElaboratorCommunication}
\setcitation{DRUP}{GN03VerificationProofsUnsatisfiabilityCNFFormulas}
\setcitation{RUP}{GN03VerificationProofsUnsatisfiabilityCNFFormulas}
\setcitation{GRIT}{CMS17EfficientCertifiedResolutionProofChecking}
\setcitation{TraceCheck}{url:TraceCheck}
\setcitation{LRAT}{CHHKS17EfficientCertifiedRATVerification}

\setcitation{PR}{HKB17ShortProofsWithoutNewVariables}
\setcitation{DRAT}{WHH14DRAT-trimEfficientCheckingTrimmingUsingExpressive}
\setcitation{satcomp2016}{BHJ17SATCompetition2016RecentDevelopments}
\setcitation{satcomp2013}{url:SATcomp2013}

\setcitation{Essence}{FHJHM08Essenceconstraintlanguagespecifyingcombinatorialproblems}
\setcitation{smodels}{SN01SmodelsSystem}
\setcitation{lparse}{S98ImplementationLocalGroundingLogicProgramsStable}
\setcitation{IDPZ3}{CVVD22IDP-Z3reasoningengineFO} %TODO replace by real publication
\setcitation{IDP-Z3}{CVVD22IDP-Z3reasoningengineFO} %} %TODO replace by real publication
\setcitation{Z3}{dB08Z3EfficientSMTSolver}

\setcitation{DMN}{DMN} %TODO is there something better than just this on-line thing?
\setcitation{Planning}{GNT04Automatedplanning-theorypractice}
\setcitation{AutomatedPlanning}{GNT04Automatedplanning-theorypractice}
\setcitation{RC2}{IMM19RC2EfficientMaxSATSolver}
\setcitation{enfragmo}{phd/Aavani14}
\setcitation{TPTP}{url:tptp}

\newcommand\citeProposition[1]{e.g.,~\cite{#1}}
\newcommand\citeEG[1]{\cite{#1}}
\begin{document}

\maketitle

\begin{abstract}
	The implicit hitting set (IHS) approach offers a general framework for solving
computationally  hard combinatorial optimization problems declaratively.
%  Constituting a hybrid approach,
IHS iterates between
a decision oracle used for
extracting sources of inconsistency
%  in the %declarative 
%  specification at hand
and an optimizer for computing so-called
hitting sets (HSs) over the accumulated sources of inconsistency.
While the decision oracle is language-specific, %to the declarative language at hand,
the optimizer is usually instantiated through  integer programming.
We explore alternative algorithmic
techniques for hitting set optimization based on different ways of employing pseudo-Boolean (PB) reasoning
as well as stochastic local search.
%The alternative HS computation approaches are
%applicable in general in different instantiations of the IHS approach.
We extensively evaluate the practical feasibility
of the alternatives in particular in the context of pseudo-Boolean (\mbox{0--1} IP) optimization
as one of the most recent instantiations of IHS.
Highlighting a trade-off between efficiency and reliability,
while a commercial IP solver turns out to remain the most effective way to instantiate
HS computations, it can cause correctness issues due to numerical instability;
%  due to potential numerical instability using an off-the-shelf IP solver as-is
%  is not generally exact and can thereby result in incorrect results;
in fact,  we show that exact HS computations instantiated via PB reasoning
can be made competitive with a numerically exact IP solver. Furthermore,
the use of PB reasoning as a basis for HS computations allows for obtaining certificates for
the correctness of IHS computations, generally applicable to any IHS instantiation in which
reasoning in the declarative language at hand can be captured in the PB-based proof format we employ.
	
	This paper is accepted for publication in the proceedings of AAAI 2026 \cite{IVSBBJ26EfficientReliableHitting-SetComputationsImplicitHitting}. 
\end{abstract}

% The table of contents below is added for your convenience. Please do not use
% the table of contents if you are preparing your paper for publication in the
% EPiC Series or Kalpa Publications series

%\setcounter{tocdepth}{2}
%{\small
%\tableofcontents}

%\section{To mention}
%
%Processing in EasyChair - number of pages.
%
%Examples of how EasyChair processes papers. Caveats (replacement of EC
%class, errors).

%------------------------------------------------------------------------------

\section{Introduction}

The implicit hitting set (IHS) approach~\cite{LS08AlgorithmsComputingMinimalUnsatisfiableSubsetsConstraints}
offers a general framework for solving                                           
computationally  hard combinatorial optimization problems declaratively.
Successful practical instantiations of IHS include solvers for
maximum satisfiability~\cite{DB11SolvingMAXSATSolvingSequenceSimplerSAT, DB13ExploitingPowermipSolversMaxSAT, DB13PostponingOptimizationSpeedUpMAXSATSolving,phd/Davies14,SBJ16LMHSSAT-IPHybridMaxSATSolver},
finite-domain constraint optimization~\cite{DB13SolvingWeightedCSPsSuccessiveRelaxations},
answer set programming~\cite{SDAJ18HybridApproachOptimizationAnswerSetProgramming}, and
pseudo-Boolean optimization~\cite{SBJ21Pseudo-BooleanOptimizationImplicitHittingSets, SBJ22ImprovementsImplicitHittingSetApproachPseudo-Boolean}, as well as
other computationally hard problems such as
computing unsatisfiable subsets of formulas~\cite{IPLM15SmallestMUSExtractionMinimalHittingSet,GBG23EfficientlyExplainingCSPsUnsatisfiableSubsetOptimization} and
quantified Boolean formulas~\cite{NMBJ22ComputingSmallestMUSesQuantifiedBooleanFormulas}, propositional abduction~\cite{IMM16PropositionalAbductionImplicitHittingSets, SWJ16ImplicitHittingSetAlgorithmsReasoningBeyond}, and
computing optimal causal graphs~\cite{ijcai/HyttinenSJ17}.

Constituting a hybrid approach, the IHS framework consists of two main components:
one for extracting sources of inconsistency in the declarative specification at hand (referred to as \emph{core extraction}, with cores being constraints ruling out sources of inconsistency),
and one for computing so-called hitting sets
over sources of inconsistency (the \emph{hitting set optimizer}).                                                 
IHS iterates between these two components, accumulating
cores and computing hitting sets over the so-far accumulated cores.
The hitting set obtained from the optimizer is used to focus the core extraction step (which reasons only on the original declarative specification) on finding sources of inconsistency that eliminate the current best candidate solution. 
%This c
%to consider only sources of inconsistency. Where as each core extraction step
%is made on the original declarative constraints restricted a hitting set without
%knowledge of the objective function at hand, 
In this way, the instance solved by the hitting set optimizer
keeps on growing as more and more cores are extracted.
The core extractor is implemented 
using a  decision oracle specific to the declarative language at hand.
In contrast,
the optimizer is quite standardly instantiated through
integer programming. 
Indeed, despite the fact that there been has significant early work put into investigating effective ways of realizing the hitting set component in IHS especially in the context of MaxSAT~\cite{DB11SolvingMAXSATSolvingSequenceSimplerSAT, DB13ExploitingPowermipSolversMaxSAT},
the use of IP solvers remains largely the main mechanism used in modern IHS implementations for efficiency and for computing  optimal solutions.	
Further improvements in the hitting set component
are generally applicable to IHS implementations.

%Focusing on the generic hitting set
%optimizer component, 
In this work
we explore alternative algorithmic
techniques for hitting set computations within IHS based, with the key aim of
realizing effective \emph{and} trustworthy hitting set computations.
In terms of algorithmic alternatives to using IP solvers,
we explore
different ways of employing pseudo-Boolean (PB) reasoning~\cite{RM09Pseudo-BooleanCardinalityConstraints}                         
as well as stochastic local search~\cite{aaai/Chu0L23,Chu0LL023} and their combinations together with IP solving.
In terms of trustworthiness, it should be noted that the hitting set optimizers is in charge
of ensuring that the objective function at hand is considered exactly.
While IP solvers remain today the de facto approach to hitting set optimization within IHS,
they are known to sometimes suffer from numerical instability, resulting in erroneous solutions \cite{CKSW13hybridbranch-and-boundapproachexactrationalmixed-integer}. 
%to aim at correct IHS implementation in terms of obtaining actual optimal solutions
%requires adjusting experimentally parameters of an IP solver in hope to avoiding
%issues with numerical instability issues intrinsic to IP solving~\cite{} and at the same
%time up-keeping the practical efficiency of the solver. 
While numerically exact
implementations of IP solvers exist~\citeEG{EG21ComputationalStatusUpdateExactRationalMixed, CKSW13hybridbranch-and-boundapproachexactrationalmixed-integer}, their performance remains today far
from the efficiency of IP solvers in their default settings without guarantees on exactness.

The alternative hitting set computation approaches are                                        
applicable in general in different instantiations of the IHS approach, and---being based
on exact reasoning---offer correctness guarantees.
We extensively evaluate the practical feasibility                                                                        
of the alternative algorithmic approaches to hitting set optimization we propose
in the context of pseudo-Boolean (0--1 IP) optimization                                 
as one of the most recent instantiations of IHS.
Highlighting a trade-off between efficiency and reliability,
while a commercial IP solver turns out to remain the most effective way to instantiate
hitting computations in this setting, due to potential numerical instability using an off-the-shelf IP solver as-is
is not generally exact;
in fact,  we show that exact HS computations instantiated via PB reasoning
can be made competitive with a numerically-exact IP solver. 
Furthermore,
the use of PB reasoning as a basis for hitting set computations allows for obtaining \emph{certificates of
	correctness} of IHS computations.  
Indeed, while not yet generally adopted in IP solvers, the idea that solvers should be \emph{certifying} (as in not only produce an answer but also a certificate of correctness of this answer), has slowly found its way from SAT solving to richer paradigms such as SMT solving  \cite{BRKLNNOP22FlexibleProofProductionIndustrial-StrengthSMTSolver}, constraint programming \cite{GMN22AuditableConstraintProgrammingSolver,FSMSD24Multi-StageProofLoggingFrameworkCertifyCorrectness} and pseudo-Boolean optimization (see, e.g., \cite{url:pbcomp2024}). 
Building on this, one of the key contributions of this paper is the development of the first \emph{certifying instantiation} of a state-of-the-art IHS
approach. 
To achieve this, we build on the VeriPB proof format~\cite{BGMN23CertifiedDominanceSymmetryBreakingCombinatorialOptimisation}, which has recently been shown to  be applicable to a rich variety of MaxSAT algorithms \cite{VDB22QMaxSATpbCertifiedMaxSATSolver,BBNOV23CertifiedCore-GuidedMaxSATSolving,BBNOPV24CertifyingWithoutLossGeneralityReasoningSolution-Improving,JBBJ25CertifyingParetoOptimalityMulti-ObjectiveMaximumSatisfiability,VCB26CertifiedBranch-and-BoundMaxSATSolving}, with IHS being the most important missing paradigm. 
Our results on certifying IHS are applicable to any IHS instantiation where the reason of the decision oracle can be captured with PB-based reasoning, and as such includes e.g.\ MaxSAT as well.

\section{Preliminaries}

%!TEX root = ../HittingSetComputationInIHS.tex 
% So that Jeremias IDE plays nice... 

\subsection{Pseudo-Boolean Optimization}

A \emph{literal}~$\litell$ is a Boolean variable $\varx$ or its negation \mbox{$\olnot{\varx} = 1 - \varx$}, where variables take values~$0$ (false) or $1$~(true).
A \emph{pseudo-Boolean (PB) constraint} $\constrC$ is a \mbox{$0$--$1$} linear inequality
$\sumnodisplay_i \coeffa_i \litell_i \geq \dega$,
where $\coeffa_i$ and~$\dega$ are integers.
Without loss of generality, we often assume PB constraints to be in \emph{normal form}, meaning that the literals $\litell_i$ are over distinct variables, each \emph{coefficient}~$\coeffa_i$ is \emph{positive} and the \emph{degree (of falsity)}~$\dega$ is non-negative. 
A \emph{pseudo-Boolean formula} $\formf$ is a conjunction $\bigand_j \constrc_j$
of PB constraints.
When convenient, we view $\formf$ as a \emph{set} of constraints. 
%For a constraint $\constrc$, $\lit(\constrc)$ is the set of literals that appear in $\constrc$.
%For a pseudo-Boolean formula $\formf$, $\lit(\formf)=\bigcup_{C\in\formf}\lit(C)$.
%
An \emph{objective} $\objective$ is an expression $\sum_i \weight_i \litell_i+\lowerbound$, where the coefficients $\weight_i$ and the constant $\lowerbound$ are integers ($\lowerbound$ stands for ``lower bound'': when the $\weight_i$ are all possible, this constant term is indeed a lower bound on the cost).

An assignment $\assmnt$ maps variables to either $0$ or $1$. The assignment $\assmnt$ satisfies a (normalized)  
PB constraint~$\constrc = \sumnodisplay_i \coeffa_i \litell_i \geq \dega$  if $\sumnodisplay_i \coeffa_i \assmnt(\litell_i) \geq \dega$ and is 
a solution to a formula $\formf$ if it satisfies all of its constraints. The cost $\assmnt(\objective)$ of an assignment $\assmnt$ under  an objective $\objective = \sum_i \weight_i \litell_i+\lowerbound$ is  $\sum_i \weight_i \assmnt(\litell_i)+\lowerbound$ (we assume without loss of generality that all the $\weight_i$ are positive).
An instance of the \emph{pseudo-Boolean optimization problem} is a tuple $(\formf,\objective)$, where $\formf$ is a PB formula and $\objective$ an objective under minimization. 
The solutions of $(\formf,\objective)$ are the solutions of $\formf$. A solution $\assmnt$ is optimal if 
$\assmnt(\objective)\leq\assmnttwo(\objective)$ for each solution $\assmnttwo$ of $(\formf,\objective)$. The optimal cost of $(\formF, \objective)$ is $\assmnt(\objective)$ for an optimal solution of $(\formF, \objective)$. 

\iffalse
\todo{check what is needed}
A \emph{substitution} (sometimes also called a \emph{witness}) $\subst$ is a mapping of variables to $0$, $1$ and literals. 
An \emph{assignment} $\assmnt$ is a substitution that maps each variable in its domain into $\{0,1\}$. 
An assignment is \emph{complete} for $\formf$ if it assigns a value to each variable of interest (where the set of variables should be clear from the context). 
We write $\subst(\constrc)$ for the constraint obtained from $\constrc$ by replacing each variable $\varx$ in the domain of $\subst$ by $\subst(\varx)$ (and implicitly normalizing);
$\subst(\objective)$, $\subst(\litell)$, $\subst(\formf)$ are defined analogously.

A (normalized)  
PB constraint~$\constrc = \sumnodisplay_i \coeffa_i \litell_i \geq \dega$
is \emph{trivial} if $\dega \leq 0$ and \emph{contradictory} if $\sum_i\coeffa_i<\dega$. 
The constraint~$\constrc$ is \emph{satisfied} by~$\assmnt$ (denoted $\assmnt\models \constrc$) if $\assmnt(\constrc)$ is trivial. 
A PB formula is satisfied if all of its constraints are satisfied. 
An instance of the \emph{pseudo-Boolean optimization problem} is a tuple $(\formf,\objective)$ with $\formf$ a PB formula and $\objective$ an objective to minimize. 
A complete assignment $\assmnt$ is a \emph{solution} of $(\formf,\objective)$ if $\assmnt\models\formf$ and is optimal if also
$\assmnt(\objective)\leq\assmnttwo(\objective)$ for each solution $\assmnttwo$ of $(\formf,\objective)$. The optimal cost of $(\formF, \objective)$ is $\assmnt(\objective)$ for an optimal solution of $(\formF, \objective)$. 
\fi 

Following~\cite{SBJ21Pseudo-BooleanOptimizationImplicitHittingSets}, a constraint $\coreconstr$ is a \emph{core} of $(\formf,\objective)$ if the following conditions hold:
(i)~the literals in $\coreconstr$ are objective literals or their negations and (ii)~all assignments $\assmnt$ that satisfy $\formF$ also satisfy $\coreconstr$ (denoted by $\formF \models \coreconstr$).
In other words, a core is an implied constraint that only mentions variables occurring in the
objective.\footnote{See Section \ref{sec:terminology} for a historic explanation of this terminology.}
%\footnote{This definition of a core comes from the first papers on IHS for PBO~\cite{SBJ21Pseudo-BooleanOptimizationImplicitHittingSets}. It should be noted
%that in the context of e.g.~Boolean satisfiability (SAT), an ``unsatisfiable core'' is standardly defined
%as a subset of clauses which are unsatisfiable.
%When phrasing optimization problems such as PBO or MaxSAT in terms of optimizing a cost function subject to (hard) constraints, cores are more conveniently defined as entailed constraints over objective variables. This is particularly
%evident in the context of PBO where cores can be more generally PB constraints than mere clausal at-least-one cardinality
%constraints as in MaxSAT, as well as IHS as the cores under our definition are the constraints which are accumulated
%in the ``hitting set'' component of IHS.}

The following proposition forms the basis for 
how the IHS approach to PBO employs cores to compute optimal solutions. 
\begin{proposition}[\citeProposition{SBJ21Pseudo-BooleanOptimizationImplicitHittingSets}]\label{prop:coreLB}
	Let $(\formF, \objective)$ be a PBO instance, $\ubModel$ an optimal solution to $(\formF, \objective)$, $\accumCores$ a set of cores of $(\formF, \objective)$, and $\csol$ an optimal solution to $(\accumCores, \objective)$. Then $\csol(\objective) \leq \ubModel(\objective)$.
\end{proposition}
In words, Proposition~\ref{prop:coreLB} states that the cost of the optimal (minimum-cost) solutions to any set of cores is a lower bound
on the optimal cost of the instance.

\subsection{The Implicit Hitting Set Approach}

Algorithm~\ref{IHS:alg} details \ihs, the implicit hitting set algorithm for pseudo-Boolean optimization~\cite{SBJ21Pseudo-BooleanOptimizationImplicitHittingSets,SBJ22ImprovementsImplicitHittingSetApproachPseudo-Boolean}. 
The high-level overview of this algorithm (when ran on an instance $(\formF,\objective)$) is as follows. 
During the execution, collect a set $\accumCores$ of cores and repeatedly solve the optimization problem $(\accumCores,\objective)$. 
When solving this problem to optimality, Proposition~\ref{prop:coreLB} guarantees that we obtain a lower bound on the optimal cost of $(\formF,\objective)$. 
Based on the outcome of such an optimization call, a decision oracle is then called under a set of assumptions (literals that are given a fixed value before search starts). 
As a result either a feasible solution (giving us an upper bound on the optimal cost) or a new core (that eliminates the previously found solution) is obtained. 
The iterations between the optimizer and the decision oracle are continued until the upper bound equals the lower bound (which is guaranteed to happen at the latest when all cores have been found).

In more detail, given a PBO instance $(\formF, \objective)$, \ihs begins by checking the existence of solutions of $(\formF, \objective)$ by invoking PB decision procedure \pb on $\formF$ (Line~\ref{l:init-s}). The call returns an indicator \textit{sat?} for satisfiability and a solution $\ubModel$ in the positive case.
Given $\ubModel$, an upper bound $\upperbound$ and a lower bound $\lowerbound$ on the optimal cost of the instance are initialized to $\ubModel(\objective)$ and $-\infty$, respectively,  on Line~\ref{l:init-e}. During search, $\ubModel$ will always contain the currently best-known solution which has the property that $\ubModel(\objective) = \upperbound$.  
A set $\accumCores$ of cores is initialized by \emph{seeding} to be the set of all constraints in $\formF$ that only mention objective literals. In other words, all constraints in  $\formF$ that are cores are immediately added to $\accumCores$.

\begin{algorithm}[!t]
	\algo{$\ihs(\formf,\objective)$}{
		\KwIn{A PBO instance $(\formf,\objective)$}
		\KwOut{An optimal solution $\ubModel$}
		$(\ubModel, \mbox{\textit{sat?}})   \gets \pb(\formf)$; \label{l:init-s} \\
		\uIf{\mbox{\textit{not sat?}}} {
			\textbf{return} ``no feasible solutions'';\label{l:early-term} 
		}
		$\upperbound \gets \ubModel(\objective)$; $\lowerbound \gets -\infty$; $\accumCores \gets \seeding(\formf)$;  \label{l:init-e}\\
		%				\hil{$\mathbf{\symmetryset \gets \compsyms(\formf)}$}; \label{l:sym-comp} \\ %%%% ONE BLUE LINE
		\While{\true}{ \label{loop-s}
			$(\csol,\lowerbound) \gets \solvehs(\accumCores, \objective, \lowerbound,\upperbound)$; \label{line:hs} \\
			\lIf{$\upperbound = \lowerbound$}{ \textbf{break}} \label{l:break1}
			$(\newCores, \assmnt) \gets \oracle(\csol, \formf,\objective)$\; \label{l:core-ex}
			\lIf{$\assmnt(\objective) < \upperbound$}{ $\upperbound \gets \assmnt(\objective)$; $\ubModel \gets \assmnt$} \label{l:updateUB}
			\lIf{$\upperbound = \lowerbound$}{ \textbf{break}\label{l:break2}}
			$\accumCores \gets \accumCores \; \cup \; \newCores$; \label{loop-e}
		}
		\textbf{return} $\ubModel$;
	}
	%			\vspace{1mm}                                                                                                                          
	\caption{IHS in the context of PBO.\label{IHS:alg}}
\end{algorithm}

The main search loop (Lines~\ref{loop-s}-\ref{loop-e}) iterates until $\lowerbound = \upperbound$, at which point $\ubModel$ is known to be optimal. 
Each iteration begins with the hitting set%\footnote{While the term ``hitting set'' is still standardly used widely in the literature on IHS, is should be noted that the constraints the hitting set component deals with are not at-least-one cardinality constraints hitting
%as in the classical hitting set problem. In particular, both seeding and the
%generality of PBO cores results in including more general PB constraints in the ``hitting set computations'' than
%mere at-least-one cardinality constraints.}
\footnotemark[\thefootnote] %Repeat previous footnotemark
component of IHS, i.e., the procedure \solvehs computing  a solution $\csol$ of the instance $(\accumCores, \objective)$ consisting of the cores found so far and the objective (Line~\ref{line:hs}). The main focus in this work is on the \solvehs procedure: we will detail various instantiations of \solvehs in Section~\ref{sec:hs-computation}. 

In addition to $\csol$, each call to \solvehs makes use of Proposition \ref{prop:coreLB} to return a lower bound $\lowerbound$ on the optimal cost of $(\formF, \objective)$.
As detailed in Section~\ref{sec:hs-computation}, the instantiations we consider will return either the input parameter $\lowerbound$ unchanged, or the cost 
 $\csol(\objective)$ when $\csol$ is known to be an optimal solution of $(\accumCores, \objective)$. 
 In a textbook implementation of plain IHS, \solvehs would always search for an optimal solution of $\accumCores$ with respect to $\objective$. 
 However, to make IHS perform well in practice, it is important to heuristically allow it to return suboptimal solutions as well. 
To guarantee termination, the heuristics must ensure a form of fairness: at any iteration of search, an optimal solution must always be returned at some later iteration. 
% In order to guarantee the termination of Algorithm~\ref{IHS:alg} with an optimal solution to $(\formF, \objective)$ we assume that 
% on each iteration $j$ there exists a $k \geq 0$ s.t. \solvehs computes an optimal solution of $(\accumCores, \objective)$ (and thus updates $\lowerbound$) on iteration 
% $j + k$. 
 For more details and a formal proof of correctness of  Algorithm~\ref{IHS:alg}, we refer the reader to~\cite{BHJS17ReducedCostFixingMaxSAT}. 

%Some of 
%instantiations of interest guarantee that $\csol$ is an optimal solution of $(\accumCores, \objective)$, which in turn by Proposition~\ref{prop:coreLB}
%implies that $\csol(\objective)$ will be a lower bound on the optimal cost of $(\formF, \objective)$. 
%Traditionally, each iteration begins with the computation of an optimal solution $\csol$ of the instance $(\accumCores, \objective)$ consisting of the cores found so far and the objective (Line~\ref{line:hs}). 
%In practice, this step is typically solved using a Mixed Integer Programming (MIP) solver.
%By Proposition~\ref{prop:coreLB}, $\csol(\objective)$ will be a lower bound on the optimal cost of the instance. 
%When solving this problem to optimality, we will hence also derive a new lower bound.
%This procedure \solvehs is the main study of interest in the current paper, where we will experiment with different possible implementations, varying in efficiency and reliability. 

% Since no cores are ever removed from $\accumCores$, the values of $\csol(\objective)$ will be non-decreasing over the iterations, so $\lowerbound$ can always be updated on Line~\ref{line:lb}. 
Next, the procedure \oracle  is used to obtain new cores. 
This is done by repeated invocations of a decision procedure for PB constraints of $\formF$ under a set of assumptions  
$\mathcal{A}$. 
This set of assumptions is initialized to the solution $\csol$, meaning that first a solution of $\formf$ that extends $\csol$ is sought for. 
If $\formF$ is unsatisfiable under $\mathcal{A}$  the decision oracle uses standard conflict analysis methods to extract a new core $C$ of $\formf$ that contradicts $\mathcal{A}$ (a core that is false whenever all literals in $\mathcal{A}$ are true) and adds it to the set $\newCores$ of new cores to be returned by \oracle and added to $\accumCores$ in Line~\ref{loop-e} of Algorithm~\ref{IHS:alg}. 
Before the next call to the decision procedure, the set $\mathcal{A}$ is relaxed by the 
so-called \emph{weight-aware core extraction}~\cite{DBLP:conf/cp/BergJ17} heuristic that---informally speaking---removes at least one of the literals in $C$ from the assumptions, thus preventing $C$ from being rediscovered in later iterations. 
The \oracle procedure terminates when the decision procedure provides a solution $\assmnt$ of $\formF \land \bigwedge_{\ell \in \mathcal{A}} \ell$. 
In case this solution has a better objective value than the currently stored best solution, the best solution is updated and termination is checked on Lines~\ref{l:updateUB} and~\ref{l:break2} of  Algorithm~\ref{IHS:alg}.

%solution $\assmnt$ that exorens 
 %assumptions $\csol$.
%invoking a decision procedure for PB constraints of $\formF$ under the assumptions $\csol$. 
%If this yields a solution, then $\csol$ can be extended to a solution of the original problem and hence, the result is an optimal solution. 
%If $\formf$ is unsatisfiable under assumptions $\csol$, a decision oracle can, using the standard conflict analysis methods, extract a new core of $\formf$ that contradicts $\csol$. 
%This core is added to $\accumCores$ in Line~\ref{loop-e}. 
%In advanced implementations, not just a single core, but multiple cores will be extracted in a single iteration. This idea is known as \emph{weight-aware core extraction} \cite{WCE}, and is the reason why our call to $\oracle$ returns a \emph{set} $\newCores$ of new cores rather than a single core. 

%\iffalse 
%\bart{I would propose to put the paragraph below back and get it out of the footnotes (now there is too much in footnotes). We can buy an extra page. Also, I think it deserves to be spelled out completely once; especially for people who do not know the terminology yet.}
% One of the reviewers was also confused about this. 

\subsection{A Remark on Terminology} \label{sec:terminology}
Much of the terminology we use  has a historic background, dating back to IHS approaches to MaxSAT \cite{DB11SolvingMAXSATSolvingSequenceSimplerSAT}. 

Firstly, our definition of ``unsatisfiable core'' differs form how the term is traditionally defined. 
Originally an ``unsatisfiable core'' was defined as a subset of the original formula that is unsatisfiable. 
In the context of assumption-based SAT solving (also employed for SAT-based optimization) fresh variables are used to represent constraints in the input formula and such a core became ``a subset of the assumptions that cannot jointly be set to a specific value'' (or alternatively, a clause over assumption variables learned by the solver). 
In the PBO setting, this was then further generalized to be any PB constraint   over assumptions (objective variables) learned by the PB solver \cite{SBJ21Pseudo-BooleanOptimizationImplicitHittingSets}. 

Secondly, this paper is about the so-called implicit hitting set approach.
The notion of ``hitting set'' has as similar origin: when all cores are clauses over objective literals, the problem solved by \solvehs is actually a minimum-cost hitting-set problem: in this case we search for a smallest-cost set of objective literals that hits each clause. 
In our more general PBO setting, however, \solvehs itself needs to solve a PBO instance, but with the special property that the constraints only mention variables occurring in the objective. 

In this paper, we opted to keep the historic terminology, even if it has become inaccurate due to the generality of the problem at hand. We do so to make the relationship with previous work on IHS for PBO \cite{SBJ21Pseudo-BooleanOptimizationImplicitHittingSets} as well as for other domains more clear.
%\fi

\section{Alternatives Approaches to Hitting Set Computations} \label{sec:hs-computation}
%!TEX root = ../HittingSetComputationInIHS.tex 
% So that Jeremias IDE plays nice... 

We now turn to investigating various alternative instantiations of the 
hitting set component (i.e., the \solvehs procedure) for computing optimal solutions to $(\accumCores, \objective)$. % consisting of the cores found so far.
%As we will see,
%there is an inherent trade-off between the efficiency of solving and the reliability of the solutions obtained. 
%Roughly speaking,
Currently efficient instantiations of \solvehs relatively standardly make use of a commercial IP solver and floating-point computations. This holds true for the current state-of-the-art PBO-IHS solver, implementing IHS for PBO, as well as state-of-the-art IHS implementations for other declarative paradigms~\cite{SBJ21Pseudo-BooleanOptimizationImplicitHittingSets,SBJ22ImprovementsImplicitHittingSetApproachPseudo-Boolean, DB13ExploitingPowermipSolversMaxSAT, DB13SolvingWeightedCSPsSuccessiveRelaxations, SDAJ18HybridApproachOptimizationAnswerSetProgramming, GBG23EfficientlyExplainingCSPsUnsatisfiableSubsetOptimization, SWJ16ImplicitHittingSetAlgorithmsReasoningBeyond,ijcai/HyttinenSJ17}.
While numerically-stable IP solvers are available today,
most IP solvers in their default settings employ floating-point computations for practical efficiency,
essentially giving up guarantees on finding optimal solutions. 
This can and has caused intrinsic problems in terms of correctness of IHS.
Towards even higher levels of correctness guarantees, verifiable proofs of optimality would be ideal.
With these motivations, we 
explore alternative instantiations of \solvehs based on recent developments in  
conflict-driven pseudo-Boolean solvers~\cite{DGDNS21CuttingCorePseudo-BooleanOptimizationCombiningCore-Guided,DGN21LearnrelaxIntegrating0-1integerlinear} that allow producing certificates of correctness of the computations 
of Algorithm~\ref{IHS:alg}, as well as various ways of combining them within the hitting set component with IP solving and stochastic local search (SLS).

\subsection{A General View on Hitting Set Solving}\label{sec:ihs-solver}
\begin{algorithm}[t]
	\algo{$\solvehs(\accumCores, \objective, \lowerbound,\upperbound)$}{
		\If{\usesls()}{\label{l:sls-check}
			$\csol \gets \sls(\accumCores, \objective)$ \label{l:sls} \\
			\If{$\csol(\objective) < \upperbound$ }{\textbf{return} $(\csol,  \lowerbound)$\label{l:sls-term}}
		}
%		\Else{
			$\text{opt?} \gets \optimalsol()$\label{l:forceopt}\\
			$\csol \gets \optimizermin(\accumCores, \objective, \upperbound, \text{opt?})$; \label{l:non-opt}\\
			\lIf{$\csol(\objective) = \upperbound \text{ \textbf{or}  opt?} $ }{\textbf{return} $(\csol,  \csol(\objective))$ \label{l:opt-term}} 
			\lElse{\textbf{return} $(\csol,\lowerbound)$\label{l:nonopt-term}}
%		}
	}
	%			\vspace{1mm}                                                                                                                          
	\caption{General view on the hitting set component, i.e., computing solutions to the cores extracted so far.\label{optimizer:alg}}
\end{algorithm}

Algorithm~\ref{optimizer:alg} details a generalized abstraction of the \solvehs procedure (the hitting set component), allowing
for different ways of combining IP solving, conflict-driven pseudo-Boolean solving, and stochastic local search.
 \ihs (Algorithm~\ref{IHS:alg}) is tasked with computing an optimal solution to an instance $(\formf, \objective)$.
It will invoke \solvehs with the set $\accumCores$ of cores extracted so far, the objective $\objective$, and the 
current upper and lower bound $(\upperbound, \lowerbound)$ on the optimal cost of $(\formf, \objective)$. 

First \solvehs invokes \usesls, a user-specified heuristic for determining whether SLS should be invoked on Line~\ref{l:sls-check}.
If the solution  $\csol$ that SLS returns has a cost lower than the current upper bound $\upperbound$, 
\solvehs terminates on Line~\ref{l:sls-term}. Otherwise, the subroutine \optimizermin is invoked on Line~\ref{l:non-opt} and tasked to 
either compute a solution $\csol$ of $(\accumCores, \objective)$ that has $\csol(\objective) < \upperbound$, or (if no such solutions exist)
an optimal solution $\csol$ of $(\accumCores, \objective)$. Since the current best solution $\ubModel$ of $(\formf, \objective)$ maintained by \ihs
has $\ubModel(\objective) = \upperbound$ and is also a solution of $(\accumCores, \objective)$, an alternative view on \optimizermin is that of
either finding a solution of $(\accumCores, \objective)$ of cost lower than $\upperbound$ or determining that such a solution does not exist. As an 
additional heuristic, we allow forcing  \optimizermin to compute an optimal solution to $(\accumCores, \objective)$, essentially forcing 
an improvement to the lower bound. More specifically, the user-specified heuristic  \optimalsol invoked on Line~\ref{l:forceopt} returns true if \optimizermin is only allowed to terminate with an optimal solution to $(\accumCores, \objective)$.

Before terminating, \solvehs checks whether the lower bound can be improved to $\csol(\objective)$ on Line~\ref{l:opt-term}. This can happen if either \optimizermin was forced to compute an optimal solution or if it was unable to find a solution that has a cost lower than $\upperbound$. Otherwise, the procedure terminates and returns 
$\csol$ and the input bound $\lowerbound$ on Line~\ref{l:nonopt-term}.

Next, we detail %the different parts of \solvehs
 different instantiations of 
\optimizermin that we consider as well as integration of local search for potential practical runtime improvements. Our general view on computing hitting sets allows for combining these instantiations in various different ways. 
	
\subsubsection{\optimizerminInTitle using an IP Solver.}
The arguably most commonly used instantiation of \optimizermin invokes 
a commercial IP solver on $(\accumCores, \objective)$. 
We consider two variants of IP solving for instantiating \optimizermin. The first variant employs fast floating-point arithmetic in its computation.
These are arguably among the most efficient methods of computing solutions $(\accumCores, \objective)$, and most state-of-the-art IHS approaches
instantiate \solvehs with a floating-point IP solver.  One disadvantage of such solvers, however, is that there have been repeated reports of such solvers outputting faulty solutions \cite{CKSW13hybridbranch-and-boundapproachexactrationalmixed-integer}, preventing their use in applications that require trust in the results.
Thus, we also consider exact IP solving, i.e., IP solving methods that employ exact arithmetic to circumvent issues 
related to floating point precision and rounding errors~\cite{CKSW13hybridbranch-and-boundapproachexactrationalmixed-integer,EG23computationalstatusupdateexactrationalmixed}.  This increased reliability of the results does, however, come at a cost in terms of computational efficiency. 
Recent computational studies~\cite{EG23computationalstatusupdateexactrationalmixed} report slowdowns of a factor $8.1$ when comparing a state-of-the-art exact IP solver to its floating point counterpart.
% Additionally.... (something about proofs?)
%\todo{unsure of what to say about proofs here, VIPR~\cite{CGS17VerifyingIntegerProgrammingResults} does not support all presolving techniques that MIP solvers do, but it does support optimization, so in theory we could get a proof for IHS with exact SCIP if we convert the VIPR proof to VeriPB, I am however unsure if we want to say this}. 
%\bart{Would propose not to talk about proofs here. Let's postpone to the proofs section}

\subsubsection{\optimizerminInTitle using Conflict-Driven PB Optimization.}\label{HS-solve-with-OPB}
Complementing the branch \& cut search commonly employed in IP solvers, we also 
consider harnessing recent developments in native pseudo-Boolean optimization procedures that reduce the problem of computing an optimal solution of $(\accumCores, \objective)$ into a sequence of decision problems. The decision problems  are tackled with a decision procedure similar to the one employed within
Algorithm~\ref{IHS:alg} to extract cores~\cite{DGDNS21CuttingCorePseudo-BooleanOptimizationCombiningCore-Guided,DGN21LearnrelaxIntegrating0-1integerlinear}. More specifically, we consider three different variants of pseudo-Boolean optimization algorithms for computing an optimal solution to $(\accumCores, \objective)$: solution-improving search (SIS), core-guided search (CG), and core-boosted search (CB). 

A SIS-based approach maintains an upper bound $\upperbound_c$ on the optimal cost of $(\accumCores, \objective)$, In the context of \solvehs, $\upperbound_c$ is instantiated to the input upper bound $\upperbound$. In each iteration, the decision procedure is invoked on $\accumCores \land (\objective < \upperbound_c)$, i.e., the cores in $\accumCores$ and a \emph{solution-improving constraint} $(\objective < \upperbound_c)$. If a solution exists, 
the returned solution $\csol$ will have $\csol(\objective) < \upperbound_c$, so the upper bound is updated, and the process repeats. Otherwise (if the decision procedure reports unsatisfiable), the latest solution found is determined to be optimal for $(\accumCores, \objective)$.

One potential issue with this SIS-based approach is that the constraints $\objective < \upperbound_c$ that are added to the solver are not sound for future calls to this oracle. Indeed, when new cores are learned, $\upperbound_c$ might no longer be an upper bound on the best objective value. 
As a consequence, a fresh instantiation of the optimization solver needs to be created each time it is called. 
To deal with this, we also consider a variant that we call reified SIS. 
An incremental SIS approach adds the reified solution-improving constraint 
\[r_{\upperbound_c} \Rightarrow (\csol(\objective) < \upperbound_c),\]
 i.e., the linear inequality 
 \[- M \olnot{r_{\upperbound_c}} + \csol(\objective) <  \upperbound_c\] where $r_{\upperbound_c}$ is a fresh variable and $M$ is a large enough constant, and invokes the decision procedure while assuming $r = 1$.
The benefit of relaxing the solution-improving constraint is that the decision procedure instantiation can be incrementally
maintained between iterations of $\solvehs$. 
Not just the constraint itself, but also all constraints derived from it will be reactivated in all future solve calls where a bound of $\upperbound_c$ or lower is enforced. 

A core-guided approach~\cite{DGDNS21CuttingCorePseudo-BooleanOptimizationCombiningCore-Guided} maintains a reformulated objective $\objective^R$, initialized to $\objective$, and a set $\mathcal{C}$ of additional constraints, initialized to $\emptyset$. In each iteration, the decision procedure is queried for a solution of $\accumCores \land \mathcal{C}$ while assuming all literals in $\objective^R$ to $0$. 
More informally, the query asks for a solution $\csol$ of $\accumCores \land \mathcal{C}$ that has $\csol(\objective^R) = 0$. If a solution exists, the obtained solution is an optimal solution of $(\accumCores, \objective)$ and the procedure terminates. Otherwise, the call returns a new core falsified by the assumptions. Next, the core-guided approach introduces new constraints to $\mathcal{C}$ and reformulates $\objective^R$ in a manner that resolves the inconsistency represented by the core in a cost-minimal way. Informally speaking, the invariant maintained by a core-guided search is that optimal solutions of $(\accumCores \land \mathcal{C}, \objective^R)$ are optimal solutions of $(\accumCores, \objective)$. Since the set $\accumCores$  of cores that we invoke core-guided search on monotonically increases during search, the state of the core-guided solver is maintained between invocations. While such \emph{incremental invocations} of core-guided search prevent fixing any variables based on the bounds on the optimal cost of $(\accumCores, \objective)$, variables are still fixed based on the bounds of the full instance $(\formF, \objective)$. For a concrete example, we fix the value of an objective literal $\ell$ that has a constant larger than $\upperbound$ to $0$ since any solution that assigns $\ell = 1$ would incur a cost larger than $\upperbound$

Finally, core-boosted search~\cite{BDS19Core-BoostedLinearSearchIncompleteMaxSAT} is a hybrid of core-guided and solution-improving search that starts with core-guided search for a user-specified number of iterations. After this, solution-improving search is invoked 
on the final reformulated instance $(\accumCores \land \mathcal{C}, \objective^R)$ obtained from the core-guided search phase. 

It is important to note that 
SIS, CG and CB can each provide intermediate solutions that allow terminating search as soon as a solution of cost lower than $\upperbound$ is found. Specifically, the instantiation of core-guided search that we use makes use of the so-called \emph{stratification} heuristic~\cite{DBLP:conf/cp/AnsoteguiBGL13} and \emph{weight-aware core extraction}~\cite{DBLP:conf/cp/BergJ17} that provide intermediate solutions, whereas  plain core-guided search does not
find any intermediate solutions before 
finding the first solution which is guaranteed to be optimal.

\subsection{Integrating Stochastic Local Search}\label{sec:sls}
Stochastic local search (SLS) is a lightweight heuristic approach to pseudo-Boolean optimization
that quickly finds low-cost solutions.
In contrast to the approaches described in Section~\ref{sec:ihs-solver}, 
SLS alone can not give any guarantee on optimality
of a solution and cannot therefore establish a lower bound on its own.
Hence an IHS approach with SLS integration in the hitting set component
cannot terminate without an instantiation of \optimizermin that
can check whether the solution found by SLS is optimal.

%Algorithm~\ref{optimizer:alg} shows our SLS integration.
%Before calling the optimizer, SLS tries to find a solution that has lower cost than
%the current upper bound.
%Whenever this call succeeds, we replaced
%an optimizer call with a, usually faster, SLS call.
%In case SLS does not find a better solution, the optimizer is called as usual.

%Thus, the benefit of SLS depends on how many costly optimizer calls it can replace,
%without incurring as little overhead as possible.
%An efficient SLS integration is further complicated by the fact that
%SLS is only beneficial when it provides solutions that guide the IHS solver
%towards an optimal solution as fast as the optimizer would.
%This is a difficult task as it is unknown what qualities in a solution are desired.
 
 Our SLS component of hitting set computations (the \sls procedure in Algorithm~\ref{optimizer:alg})
 is designed to achieve two desirable properties of SLS search in the context of \solvehs. 
 First, the search should incur as little overhead as possible to justify 
 replacing calls to \optimizermin that can result in improved bounds on the optimal cost. 
 Second, the solutions found by SLS should, intuitively, be as helpful as those 
 found by \optimizermin in guiding IHS search toward an optimal solution of the full instance.

Our instantiation of \sls is built on recent advances in SLS for
pseudo-Boolean optimization~\cite{Chu0LL023} and adapts the NuPBO approach to the
hitting set context.
NuPBO starts from a complete assignment that is not necessarily a solution.
In each iteration, the search flips the value of one variable in the hopes of eventually finding a solution.
The choice of variable is based on a score that considers three aspects:
(i) the contribution of the current variable value to satisfying the cores,
(ii) how much flipping the variable value would contribute towards satisfying the  cores,
and (iii) how flipping the variable value would impact the cost of the assignment.

In more detail, we
design \sls  for incremental calls, to run in conjunction with another instantiation of \optimizermin,
and to produce good solutions for core extraction.
Adapting the SLS approach to incremental calls is straightforward, as we can keep the state
from the previous call and simply add the new constraints representing additional cores.
Effectively, \sls seeks to repair the previous solution to satisfy the new constraints.
and (iii) how flipping the variable value would impact the cost of the assignment.

%\bart{I don't understand the paragraph that follows.
%        How does \sls require an optimal solver? The local searhc part does not, right, it just does local search.                                           And then there are cases in which we
%         - either do not call sls (if the heuristics use-sls is false)
%         - or sls does not find a good enoug hsolution to continue with (line 7-8), maybe for timeout or whatever...
%         Are these the cases that are described here? SHould be made inline with pseudocode
%        }

The solution returned by \solvehs is used to extract the next set of cores.
Thus it is important that the solutions returned by \solvehs are not
only of low cost, but also guide the IHS solver towards an optimal solution.
We incorporate this idea in our SLS integration by ensuring that the solutions
are \emph{diversified}, i.e., that the subsequent solutions returned by \solvehs
have a large Hamming-distance, whenever possible.
Hence, we split \sls into two phases: (i) we start from the previous solution
and (ii) we restart the search and start from the previous solution but randomly
flip a large number of variable values.
Whenever we find a solution in both phases, we prefer the one of lower cost. In case
of a tie, we prefer the solution from the second phase to foster diversification.
Further, \sls's purpose is avoiding costly calls to the optimizer.
Hence we skip \sls whenever we know from prior iterations
that the optimizer calls do not take longer than \sls calls.

%\bart{Which one do you keep? THe best of the two?
%        Do you ALXWAYS run it twice or flip a coin which one to run?
%        Also to be made more inline with the pseudocode above. \sls is only called once in Algorithm2.
%        \textit{inside} sls you might be running two local search iterations and then do something with the two results (rewrite or update pseudo-code)}

%
%Since SLS approaches cannot, in general, determine the optimality of solutions that they compute,
%SLS alone is not sufficient whenever \solvehs needs to compute an optimal solution
%(i,e. when the opt-sol boolean in Algorithm~\ref{optimizer:alg}
%is true).
%We do, however, still invoke SLS in those cases as well and use the solution computed to warmstart the instantiation of \optimizermin invoked afterward. \todo{feel free to change this if this is not true}
%\bart{Really? Hmm I reordered the algorithm earlier to do the opt-sol bit FIRST and only the nthe SLS bit, because
%        a) it seemed useless to call SLS if we are looking for an optimal solution anyway
%        b) the pseudocode completely ignored the solution SLS found so there was no point in having it there.
%        To be checked!!}

\section{Certified IHS Computations} \label{sec:proofs}
%!TEX root = ../HittingSetComputationInIHS.tex 
% So that Jeremias IDE plays nice.

We now turn to developing \emph{proofs}, i.e., independently checkable certificates of correctness of results, 
for the IHS paradigm.
We employ the \veripb format as a general-purpose proof format for pseudo-Boolean optimization, 
combining implicational reasoning using cutting planes operations (such as adding up two PB constraints) \mycite{CuttingPlanes} with well-chosen strengthening rules that allow deriving non-implied constraints \cite{BGMN23CertifiedDominanceSymmetryBreakingCombinatorialOptimisation}. 
For representational succinctness of the current paper, details of the actual proof rules are not critical to spell out, but it is relevant that the strengthening rules allow us to derive fresh reification variables.
In other words, if $x$ is a variable that does not yet appear in the proof and $C$ is a PB constraint, we can always derive (one or both) two new PB constraints $x\Rightarrow C$ and $x \Leftarrow C$ that express that $x$ is true precisely when $C$ is satisfied. 
If $C$ is the normalized constraint $\sumnodisplay_i \coeffa_i \litell_i \geq \dega$, then $x\Rightarrow C$ and $x \Leftarrow C$ are 
\begin{align*}
	&\dega \olnot x + \sumnodisplay_i \coeffa_i \litell_i \geq \dega \text{ and }\\
	& (\sumnodisplay_i \coeffa_i -\dega + 1)x + \sumnodisplay_i \coeffa_i \olnot \litell_i \geq \sumnodisplay_i \coeffa_i -\dega + 1, 
\end{align*}
respectively.

%\subsubsection{Preliminaries: The VeriPB Proof Format} 

\subsection{VeriPB Proofs for \ihsInTitle}
Pseudo-Boolean solvers using solution-improving search and core guided search have had support for \veripb proof logging for a while (see, e.g., the recent
certifying track in the PBO Competitions \cite{url:pbcomp2024}). 
As such, a natural idea toward 
obtaining uniform proofs for \ihs is to instantiate \solvehs with a PB-based approach and to combine the proofs produced by the \solvehs subroutine with the 
proofs produced by the \oracle subroutine. While the idea is conceptually simple, there are a number of details that need to be 
considered. 
%If we use an PBO solver for the \solvehs algorithm (as detailed in Section~\ref{HS-solve-with-OPB}), then we can harness the fact that these solvers have had support %for \veripb proof logging for a while (see, e.g., the certifying track in the PBO competition \cite{}). 
%In this setting, given the fact that both the decision oracle \pb and the optimizer used for \solvehs can produce \veripb proofs, it is to be expected that these can be %combined into a single holistic proof of the entire IHS pipeline. 
%In principle, this indeed works. 
Due to space restrictions, we do not go into detail about how the proof production for the different components works. Instead, we discuss challenges in their integration and how they are addressed.

\paragraph{Constraint ID and Variable Management.}
With two different solvers co-producing parts of the proof, neither of the solvers are fully aware of which constraints are derived in the proofs and, in case they introduce new variables, which variables are unused so far. 
To facilitate this, we developed a general pseudo-Boolean proof logging API which both solvers use and which is responsible for maintaining relevant information, such as mappings from solver-specific variables to their name used in the proof. 
We expect this API to be useful more generally for future pseudo-Boolean proof logging approaches as well. 

\paragraph{Non-Reified Solution-Improving Constraints.}
Another issue that needs to be addressed is that the optimizer uses (\solvehs) is not optimizing the original problem, but a derived problem.  
When using a pseudo-Boolean optimizer with proof logging, it will log solutions it finds and might then use the associated solution-improving constraint expresses that from now on we are only interested in better solutions. 
Taking a holistic perspective, however, this solution-improving constraint is not sound: the solution \solvehs finds is not a solution of the whole problem. 
We already described in Section~3 two versions of integrating a SIS solver (incremental and non-incremental). 
In the incremental case the solver does not introduce a proper solution-improving constraint but a reified version thereof; this is easy to deal with from a proof logging perspective. 
%\todo{One reviewer: Include a compact example illustrating how solution-improving constraints are reified for the proof logger and how constraint/variable namespaces are coordinated. }
In the non-incremental version, however, the solver is ran without assumptions and restarted upon every call. 
In this case the solver will be deriving those unsound solution-improving constraints. 
Our solution for dealing with this is to allow for a small discrepancy between the constraints the solver sees and the constraints derived in the proof, ensuring that the solution-improving constraint logged in the proof is indeed reified. 

\begin{example}
%	{\color{red}UNFINISHED EXAMPLE}
	Assume that the objective at hand is $x_1+ 2x_2+3x_3$ and during an execution of \solvehs we find a solution with $x_1 = x_2 =1$ and $x_3=0$. 
	While this is a solution of $\accumCores$, we do not know if this can be extended to a solution of $\formF$. 
	Deriving the solution-improving constraint 
	\begin{align} x_1+ 2x_2+3x_3 &< 3\\
		\text{(which normalizes to } \olnot x_1 + 2\olnot x_2 +3\olnot x_3 &\geq 4\text)\label{constr:solver}
		\end{align}
	in the proof would be unsound. 
	In reified SIS this is not a problem since this constraint is not learned unconditionally. 
	In regular SIS, however, when adding proof logging, we do not wish to change the behaviour of the solver.
	To this end, we add the above constraint to the solver but in the proof instead derive the reification constraint 
	\begin{align}4 \olnot r_3 + \olnot x_1 + 2\olnot x_2 +3\olnot x_3 &\geq 4\label{constr:proof}\end{align}
	Later on, whenever the solver uses constraint  \eqref{constr:solver} for deriving new constraints internally, the proof will instead reference constraint \eqref{constr:proof}, resulting in all derivations of it to be conditional on $r_3$. 
	If  it later turns out that there is solution of cost $3$ for the original PBO instance, we can derive without loss of generality that $r_3$ is indeed true. 
	
\end{example}

%\paragraph{Core Boosting}
% When doing core boosting, we do objective rewrites. Needs to be translated back to the original objective. 

\subsection{Proofs and IP Instantiations of \solvehsInTitle} 
To harness both the effectiveness of SLS and floating-point IP solvers as well as proof-production capabilities of pseudo-Boolean instantiations of \solvehs, we outline hybrid instantiations of \optimizermin that combine these strengths. The intuition here is that no proofs are needed for the calls 
to \solvehs that do not result in optimal solutions of $(\accumCores, \objective)$. The hybrid schemes  we consider seek to minimize the calls to the (less efficient) proof-producing optimizers, while still enabling proofs for the overall IHS algorithm. 
More specifically, we consider the following hybridizations of non-proof-producing (i.e., SLS or IP) instantiations of \optimizermin, and the proof-producing ones. 

\paragraph{Verify optimal solution (OptLB).} The first approach invokes a proof-producing optimizer whenever the lower bound would be increased to equal the upper bound, i.e., when an optimal solution to the overall instance has been found. Intuitively, this represents the minimum amount of exact computation in \solvehs needed to obtain a proof of correctness for the overall IHS algorithm. However, no guarantees on the intermediate lower bounds are obtained. Additionally, re-invoking a proof-producing optimizer when the LB would be increased might require some amount of duplicate work in the optimizer recomputing a solution already found by the IP or SLS solver.

\paragraph{Verify all lower bounds (AllLB).} The second approach uses a proof-producing optimizer whenever the lower bound would be refined, i.e., 
whenever the if statement on Line~\ref{l:opt-term} of Algorithm~\ref{optimizer:alg} is true. Verifying all of the lower bounds requires more (potentially duplicated) exact computation. Nevertheless, this allows for generating proofs for more search heuristics used by instantiations of IHS that we are aware of.

\paragraph{Switch when forced lower bounds (ForceLB).} The previous two hybridizations possibly perform duplicate work (asking a proof-producing optimizer to confirm all/some optimality claims of other optimizers), which might be undesirable.
In this third hybridization, we make an a priori distinction between calls for which optimality is required and calls for which optimality is not required.
Specifically, this version uses proof-producing optimizers to \emph{only} instantiate the calls to \optimizermin in which the parameter opt? is true.
The calls to \optimizermin that are not required to find an optimal solution can be instantiated with non-proof producing optimizers. 
This ensures that any calls to \solvehs that are required to find an optimal solution are indeed proof-producing. 
In this setting, the case $\csol(\objective) = \upperbound$ of the if statement on  Line~\ref{l:opt-term} that represents 
\optimizermin returning an optimal solution even when opt? is false, is not used since non-proof-producing optimizers may make such calls. 

% any optimal solutions returned by \optimizersol cannot be used to improve the lower bounds. \todo[inline]{details here. What do we actually do?}
%\todo{Bart: should not be called ``VERIFY''. Verify is after the fact for me}

\iffalse
\section{Realizing Hitting Set Computations}
\input{implementation}
\fi

\section{Experiments}
%!TEX root = ../HittingSetComputationInIHS.tex 
% So that Jeremias IDE plays nice...

In our experiments, we evaluate the performance of the different instantiations
of \optimizermin as well as the overhead caused by proof logging.
First, we compare how different solvers used as the single solver for \optimizermin
perform as well as the hybrid configurations outlined 
in Section~\ref{sec:proofs}.
Second, we evaluate the benefits of integrating SLS. Finally, we investigate
the impact of proof logging on runtime performance.
All implementation code, together with run scripts and raw data, can be found online \cite{IVSBBJ25Zenodo}.

\paragraph{Setup}
The experiments reported on were run on compute nodes with two AMD EPYC 7513 CPUs, each with 64 cores running at 2.6 GHz.
We reduced runtime variance by reserving whole CPUs for our experiments and run 16 instances in parallel on one CPU.
Each run is limited to 14 GB memory and 1 h runtime using
runlim\footnote{\url{https://github.com/arminbiere/runlim}}, and each solver is executed once on each instance.

\paragraph{Solvers}
All our implementation builds on the PBHS solver.\footnote{\url{https://bitbucket.org/coreo-group/pbhs}}
 We use Gurobi~11.0.3 and SCIP~9.2.3\footnote{\url{https://github.com/scipopt/scip} (commit~3045f20)}~\cite{BolusaniEtal2024OO} as IP solvers.
The latter offers an exact mode (see Section~\ref{sec:ihs-solver}).
Further, we use a modified version of Roundingsat\footnote{\url{https://gitlab.com/MIAOresearch/software/roundingsat} (Commit: c548e109)}~\cite{GN21CertifyingParityReasoningEfficientlyUsingPseudo-Boolean,BGMN23CertifiedDominanceSymmetryBreakingCombinatorialOptimisation}
as the proof producing PB solver and VeriPB~2.2.2\footnote{\url{https://gitlab.com/MIAOresearch/software/VeriPB}}
to check these proofs. Proof logging is enabled only for experiments concerning proof logging.

\paragraph{Instances}
We evaluate the solvers on the same set of  $1786$ instances  across tens of different benchmark domains as used in the
the original papers presenting PBO-IHS~\cite{SBJ21Pseudo-BooleanOptimizationImplicitHittingSets,SBJ22ImprovementsImplicitHittingSetApproachPseudo-Boolean}; see~\cite{SmirnovMSc} for details.

\subsection{Impact of Hitting Set Oracles} 

\begin{table}[!t]
    \centering
	\small
    \begin{tabular}{@{}l@{\hspace{10pt}}
	r@{\hspace{6pt}}r@{\hspace{10pt}}
	r@{\hspace{6pt}}r@{\hspace{10pt}}
	r@{\hspace{6pt}}r@{\hspace{10pt}}
	r@{\hspace{6pt}}r@{}}
        \toprule
		&&&\multicolumn{6}{c}{Hybrid}\\
		\cmidrule{4-9}
		&\multicolumn{2}{c}{Single}&\multicolumn{2}{c}{AllLB}&\multicolumn{2}{c}{ForceLB}&\multicolumn{2}{c}{OptLB}\\
		Solver & \# & m  & \# & m & \# & m & \# & m \\
		\cmidrule(r){1-1}
		\cmidrule(r){2-3}
		\cmidrule(r){4-5}
		\cmidrule(r){6-7}
		\cmidrule{8-9}
%		\midrule
		Gurobi&951&0.66&-&-&-&-&-&-\\
SCIP&913&1.16&-&-&-&-&-&-\\
Exact SCIP&762&2.10&-&-&-&-&-&-\\
		\cmidrule(r){1-1}
		\cmidrule(r){2-3}
		\cmidrule(r){4-5}
		\cmidrule(r){6-7}
		\cmidrule{8-9}
Core Guided&687&0.63&730&0.74&725&0.72&729&0.73\\
SIS Reified&651&1.52&654&1.41&648&1.59&704&1.22\\
SIS&672&0.79&684&0.86&697&0.88&696&0.80\\
SIS + CB&716&2.62&737&1.55&731&2.08&746&1.46\\
		\bottomrule
    \end{tabular}
	\caption{Different instantiations of \optimizerminInTitle.
		\emph{Single} shows instantiations using only a single solver and \emph{Hybrid}
		shows instantations that combine an IP solver (Gurobi) with a proof producing
		solver (Roundingsat).
		\#: number of instances solved, m: average memory usage
		in GB.}
    \label{tab:results-solvers}
\end{table}

In the first experiment, we use a single solver for the hitting set computation
in \optimizermin.
Table~\ref{tab:results-solvers} shows the results for different solvers, split
into IP solvers (top) and designated PB solving methods.
These results show that the IP solvers solve more instances.
However, the difference between SCIP and Exact SCIP show that guaranteed optimality
causes a significant performance decrease.
Indeed, the number of solved instances decreases by more than 15\% while the memory
usage doubles.

As far as the dedicated PBO methods are concerned, first we notice that the variant of solution-improving search with reified solution-improving constraints appears to suffer from too much memory usage and is outperformed by pure SIS. 
This difference disappears when using the OptLB hybrid strategy and the number of calls to the PBO oracle is significantly decreased. 
Over the line, we observe that the combination of solution-improving search with core-boosting outperforms the SIS as well as the core-guided implementation. 

The dedicated PBO methods solve fewer instances than the IP solvers, but they do come with certification, motivating the hybrid
approaches.
The more we delay the call to RoundingSat, the better the results:
AllLB has the most overhead and performs overall poorest, while OptLB requires the
fewest RoundingSat calls and performs best overall.
%Surprisingly, SIS with core boosting does not perform well on its own, being
%the second-worst method on its own, but performs best in our hybrid approaches.
%The difference might be due to an overhead in core boosting that only
%affects performance in frequent incremental calls.

%\subsection{Combinations of Hitting Set Solvers}

\subsection{Impact of Integrating Local Search}

\begin{table}[!b]
    \centering
    \begin{tabular}{lrrrrrr}
		\toprule
		&\multicolumn{3}{c}{IHS + SLS} & \multicolumn{3}{c}{OptLB + SLS}\\
		Solver & \# & m & d & \# & m &d \\
		\cmidrule(r){1-1}
		\cmidrule(r){2-4}
		\cmidrule{5-7}
		Gurobi&+17&-0.02&31&-&-&-\\
SCIP&+4&-0.14&18&-&-&-\\
Exact SCIP&+44&-0.48&58&-&-&-\\
		\cmidrule(r){1-1}
		\cmidrule(r){2-4}
		\cmidrule{5-7}
Core Guided&+41&+0.02&53&+11&-0.02&21\\
SIS Reified&+10&-0.06&52&+7&-0.04&31\\
SIS&+13&+0.02&39&+7&-0.02&25\\
SIS + CB&+19&-0.53&53&+3&-0.03&27\\
\bottomrule
    \end{tabular}
	\caption{Different instantiations of \optimizerminInTitle with and without SLS.
        \#: number of solved instances (\#); m: average memory usage in GB 
                (positive numbers indicate an increase with SLS);
        d:  number of instances solved with or without SLS but not by both configurations.}
    \label{tab:results-ls}
\end{table}

Secondly, we integrated \sls (cf.~Section~\ref{sec:sls}) into the single solver
instantiations of \optimizermin and the best hybrid approach OptLB.
By Table~\ref{tab:results-ls},
integration of SLS  improves  the runtime performance of each of the instantiations.
We use two SLS configurations, one for SLS combined with an IP solver and one for
SLS combined with dedicated PBO approaches.
In the hybrid configuration, we use the SLS configuration for IP.
Integration of SLS consistently results in more instances solved,
decrease  memory usage and improve runtimes.
This is most striking for Exact SCIP, for which SLS gives the biggest gain
in the number of solved instances, also decreasing the memory consumption
by almost 20\%.
Interestingly, often
large numbers of instances are solved either with or without SLS but not both (Column~d
in Table~\ref{tab:results-ls}).
However, when SLS does not find a sufficiently low-cost solution it directly incurs
a runtime overhead.
Despite diversification, SLS can get stuck in local optima and 
cannot escape it without an optimizer. This suggests that
improving SLS heuristics further for the IHS use case has potential
for even further improvements.

In the hybrids, SLS can only replace IP solver calls
due to inexactness.
Here, SLS contributions significantly decrease, as the number of instances
that SLS performs well on is balanced with instances it does poorly on.
It is unlikely that  instances the  IP solver can only solve when SLS is used could
be solved by the dedicated PBO approach.

%Next, we evaluate how proof logging impacts the solving process.

\subsection{Proof Logging Overhead}
We evaluate proof logging using the most performant proof-producing
variant, OptLB using SIS + CB.
%Since \sls did not significant improvements here, we run this experiment without SLS.
The proofs are checked using VeriPB with a timelimit of 10 h.
By proof logging, we found several occasions where the IP solver
wrongly claimed optimality;
Gurobi on four instances a total of nine times,
and four wrong results for Exact SCIP.

\begin{figure}[!t]
	\centering
	\includegraphics[scale=0.9]{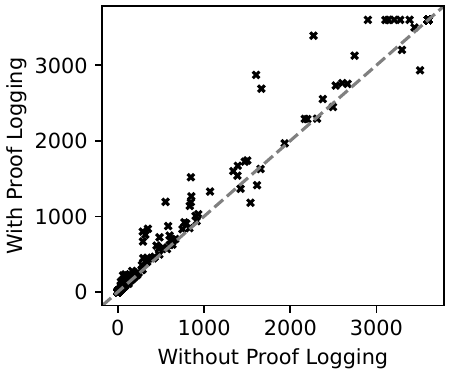}
	\caption{Comparison of runtime with and without proof logging. Times are in seconds.}
	\label{fig:pl-runtimes}
\end{figure}

SIS + CB solved 746 instances without proof logging and 741 with proof logging.
The median runtime overhead caused by proof logging is very low, 10.7\%, while the
overhead at the 90th percentile is 100.0\% (see Figure~\ref{fig:pl-runtimes}).
Although beyond the scope of this paper's contributions, with the present version of the VeriPB proof checker,
the median proof \emph{checking} time
is 588.0\% of solving with proof logging;
%  and the overhead at the 90-th percentile is 2178.3\%.
712 proofs could be checked with VeriPB with a 10 h time limit.
We expect these numbers to decrease significantly with on-going advances in VeriPB proof checking~\cite{PBOxide} once extended to optimization.

%We define the checking overhead similarly (see also Figure~\ref{fig:pl-runtimes}).

\section{Conclusion}
%The implicit hitting set approach offers a generic framework for
%developing practical solvers for NP-hard optimization problems,
%with successful instantiations for various declarative languages.
%While the specifics of the
%core extraction component of IHS is dependend on the declarative
%language at hand,
The hitting set component in IHS is standardly realized
using (typically commercial) IP solvers using floating-point computations.
Howver, this can lead to correctness issues in practice due to numerical
issues and at least at present hinders gaining trustworthy IHS computations
due to difficulties in uniformly certifying the computations of both
the core extraction and the hitting set components.
With these motivations, focusing on IHS for PBO as one of the most
recent successful IHS instantiations, we explored different combinations of
recent advances in exact PB reasoning, local search and IP solving for instantiating
the hitting set component as potential
means of gaining trustworthiness to IHS computations, addressing both
issues with numerical instability and the challenge of realizing certified IHS.
We extensively evaluate a wide range of different variants of such combinations,
and showed that by clever combinations of IP solving, PB reasoning and local search,
certified IHS can  be realized and uniform proofs obtained for IHS computations
while remaining reasonably performant, especially
when compared to employing numerically-exact IP solving.
The ideas presented are expected to be applicable to various other
IHS instantiations, such as state-of-the-art IHS-based solvers for MaxSAT, which
is a promising direction for further work.

\section*{Acknowledgments}
This work is partially funded by the European Union (ERC, CertiFOX, 101122653). Views and opinions expressed are however those of the author(s) only and do not necessarily reflect those of the European Union or the European Research Council. Neither the European Union nor the granting authority can be held responsible for them.

In addition, this work is partially funded by the Fonds Wetenschappelijk Onderzoek -- Vlaanderen (G064925N and G070521N \& fellowship 11A5J25N)
and by the Research Council of Finland (grants 362987 and 356046), and by an Amazon Research Award (Fall/2023).

The authors acknowledge support by the state of Baden-Württemberg through bwHPC
and the German Research Foundation (DFG) through grant INST 35/1597-1 FUGG.

%%------------------------------------------------------------------------------
%\section{Acknowledgments}
%\label{sect:acks}

\newcommand{\etalchar}[1]{$^{#1}$}


\begin{thebibliography}{CCL{\etalchar{+}}23b}

\bibitem[ABGL13]{DBLP:conf/cp/AnsoteguiBGL13}
Carlos Ans{\'{o}}tegui, Maria~Luisa Bonet, Joel Gab{\`{a}}s, and Jordi Levy.
\newblock Improving {WPM2} for (weighted) partial maxsat.
\newblock In {\em {CP}}, volume 8124 of {\em Lecture Notes in Computer
  Science}, pages 117--132. Springer, 2013.

\bibitem[BBB{\etalchar{+}}24]{BolusaniEtal2024OO}
Suresh Bolusani, Mathieu Besan{\c{c}}on, Ksenia Bestuzheva, Antonia Chmiela,
  Jo{\~{a}}o Dion{\'{i}}sio, Tim Donkiewicz, Jasper van Doornmalen, Leon
  Eifler, Mohammed Ghannam, Ambros Gleixner, Christoph Graczyk, Katrin Halbig,
  Ivo Hedtke, Alexander Hoen, Christopher Hojny, Rolf van~der Hulst, Dominik
  Kamp, Thorsten Koch, Kevin Kofler, Jurgen Lentz, Julian Manns, Gioni Mexi,
  Erik M\"{u}hmer, Marc~E. Pfetsch, Franziska Schl{\"o}sser, Felipe Serrano,
  Yuji Shinano, Mark Turner, Stefan Vigerske, Dieter Weninger, and Lixing Xu.
\newblock {The SCIP Optimization Suite 9.0}.
\newblock Technical report, Optimization Online, February 2024.

\bibitem[BBN{\etalchar{+}}23]{BBNOV23CertifiedCore-GuidedMaxSATSolving}
Jeremias Berg, Bart Bogaerts, Jakob Nordstr{\"{o}}m, Andy Oertel, and Dieter
  Vandesande.
\newblock Certified core-guided {MaxSAT} solving.
\newblock In {\em {CADE}}, volume 14132 of {\em {LNCS}}, pages 1--22. Springer,
  2023.

\bibitem[BBN{\etalchar{+}}24]{BBNOPV24CertifyingWithoutLossGeneralityReasoningSolution-Improving}
Jeremias Berg, Bart Bogaerts, Jakob Nordstr{\"{o}}m, Andy Oertel, Tobias
  Paxian, and Dieter Vandesande.
\newblock Certifying without loss of generality reasoning in solution-improving
  maximum satisfiability.
\newblock In {\em {CP}}, volume 307 of {\em LIPIcs}, pages 4:1--4:28. Schloss
  Dagstuhl, 2024.

\bibitem[BDS19]{BDS19Core-BoostedLinearSearchIncompleteMaxSAT}
Jeremias Berg, Emir Demirovic, and Peter~J. Stuckey.
\newblock Core-boosted linear search for incomplete {MaxSAT}.
\newblock In {\em {CPAIOR}}, volume 11494 of {\em {LNCS}}, pages 39--56.
  Springer, 2019.

\bibitem[BGMN23]{BGMN23CertifiedDominanceSymmetryBreakingCombinatorialOptimisation}
Bart Bogaerts, Stephan Gocht, Ciaran McCreesh, and Jakob Nordstr{\"{o}}m.
\newblock Certified dominance and symmetry breaking for combinatorial
  optimisation.
\newblock {\em J. Artif. Intell. Res.}, 77:1539--1589, 2023.

\bibitem[BHJS17]{BHJS17ReducedCostFixingMaxSAT}
Fahiem Bacchus, Antti Hyttinen, Matti J{\"{a}}rvisalo, and Paul Saikko.
\newblock Reduced cost fixing in {MaxSAT}.
\newblock In {\em {CP}}, volume 10416 of {\em {LNCS}}, pages 641--651.
  Springer, 2017.

\bibitem[BJ17]{DBLP:conf/cp/BergJ17}
Jeremias Berg and Matti J{\"{a}}rvisalo.
\newblock Weight-aware core extraction in sat-based maxsat solving.
\newblock In {\em {CP}}, volume 10416 of {\em Lecture Notes in Computer
  Science}, pages 652--670. Springer, 2017.

\bibitem[BRK{\etalchar{+}}22]{BRKLNNOP22FlexibleProofProductionIndustrial-StrengthSMTSolver}
Haniel Barbosa, Andrew Reynolds, Gereon Kremer, Hanna Lachnitt, Aina Niemetz,
  Andres N{\"{o}}tzli, Alex Ozdemir, Mathias Preiner, Arjun Viswanathan, Scott
  Viteri, Yoni Zohar, Cesare Tinelli, and Clark~W. Barrett.
\newblock Flexible proof production in an industrial-strength {SMT} solver.
\newblock In {\em {IJCAR}}, volume 13385 of {\em {LNCS}}, pages 15--35.
  Springer, 2022.

\bibitem[CCL23a]{aaai/Chu0L23}
Yi~Chu, Shaowei Cai, and Chuan Luo.
\newblock Nuwls: Improving local search for (weighted) partial maxsat by new
  weighting techniques.
\newblock In {\em {AAAI}}, pages 3915--3923. {AAAI} Press, 2023.

\bibitem[CCL{\etalchar{+}}23b]{Chu0LL023}
Yi~Chu, Shaowei Cai, Chuan Luo, Zhendong Lei, and Cong Peng.
\newblock Towards more efficient local search for pseudo-boolean optimization.
\newblock In Roland H.~C. Yap, editor, {\em 29th International Conference on
  Principles and Practice of Constraint Programming, {CP} 2023, August 27-31,
  2023, Toronto, Canada}, volume 280 of {\em LIPIcs}, pages 12:1--12:18.
  Schloss Dagstuhl - Leibniz-Zentrum f{\"{u}}r Informatik, 2023.

\bibitem[CCT87]{CCT87complexitycutting-planeproofs}
William~J. Cook, Collette~R. Coullard, and Gy{\"{o}}rgy Tur{\'{a}}n.
\newblock On the complexity of cutting-plane proofs.
\newblock {\em Discret. Appl. Math.}, 18(1):25--38, 1987.

\bibitem[CKSW13]{CKSW13hybridbranch-and-boundapproachexactrationalmixed-integer}
William~J. Cook, Thorsten Koch, Daniel~E. Steffy, and Kati Wolter.
\newblock A hybrid branch-and-bound approach for exact rational mixed-integer
  programming.
\newblock {\em Math. Program. Comput.}, 5(3):305--344, 2013.

\bibitem[Dav14]{phd/Davies14}
Jessica Davies.
\newblock {\em Solving {MAXSAT} by Decoupling Optimization and Satisfaction}.
\newblock PhD thesis, University of Toronto, Canada, 2014.

\bibitem[DB11]{DB11SolvingMAXSATSolvingSequenceSimplerSAT}
Jessica Davies and Fahiem Bacchus.
\newblock Solving {MAXSAT} by solving a sequence of simpler {SAT} instances.
\newblock In {\em {CP}}, volume 6876 of {\em {LNCS}}, pages 225--239. Springer,
  2011.

\bibitem[DB13a]{DB13ExploitingPowermipSolversMaxSAT}
Jessica Davies and Fahiem Bacchus.
\newblock Exploiting the power of mip solvers in {MaxSAT}.
\newblock In {\em {SAT}}, volume 7962 of {\em {LNCS}}, pages 166--181.
  Springer, 2013.

\bibitem[DB13b]{DB13PostponingOptimizationSpeedUpMAXSATSolving}
Jessica Davies and Fahiem Bacchus.
\newblock Postponing optimization to speed up {MAXSAT} solving.
\newblock In {\em {CP}}, volume 8124 of {\em {LNCS}}, pages 247--262. Springer,
  2013.

\bibitem[DB13c]{DB13SolvingWeightedCSPsSuccessiveRelaxations}
Erin Delisle and Fahiem Bacchus.
\newblock Solving weighted {CSP}s by successive relaxations.
\newblock In {\em {CP}}, volume 8124 of {\em {LNCS}}, pages 273--281. Springer,
  2013.

\bibitem[DGD{\etalchar{+}}21]{DGDNS21CuttingCorePseudo-BooleanOptimizationCombiningCore-Guided}
Jo~Devriendt, Stephan Gocht, Emir Demirovic, Jakob Nordstr{\"{o}}m, and
  Peter~J. Stuckey.
\newblock Cutting to the core of pseudo-{Boolean} optimization: Combining
  core-guided search with cutting planes reasoning.
\newblock In {\em {AAAI}}, pages 3750--3758. {AAAI} Press, 2021.

\bibitem[DGN21]{DGN21LearnrelaxIntegrating0-1integerlinear}
Jo~Devriendt, Ambros~M. Gleixner, and Jakob Nordstr{\"{o}}m.
\newblock Learn to relax: Integrating 0-1 integer linear programming with
  pseudo-{Boolean} conflict-driven search.
\newblock {\em Constraints An Int. J.}, 26(1):26--55, 2021.

\bibitem[EG21]{EG21ComputationalStatusUpdateExactRationalMixed}
Leon Eifler and Ambros~M. Gleixner.
\newblock A computational status update for exact rational mixed integer
  programming.
\newblock In {\em {IPCO}}, volume 12707 of {\em {LNCS}}, pages 163--177.
  Springer, 2021.

\bibitem[EG23]{EG23computationalstatusupdateexactrationalmixed}
Leon Eifler and Ambros~M. Gleixner.
\newblock A computational status update for exact rational mixed integer
  programming.
\newblock {\em Math. Program.}, 197(2):793--812, 2023.

\bibitem[FSM{\etalchar{+}}24]{FSMSD24Multi-StageProofLoggingFrameworkCertifyCorrectness}
Maarten Flippo, Konstantin Sidorov, Imko Marijnissen, Jeff Smits, and Emir
  Demirovic.
\newblock A multi-stage proof logging framework to certify the correctness of
  {CP} solvers.
\newblock In {\em {CP}}, volume 307 of {\em LIPIcs}, pages 11:1--11:20. Schloss
  Dagstuhl, 2024.

\bibitem[GBG23]{GBG23EfficientlyExplainingCSPsUnsatisfiableSubsetOptimization}
Emilio Gamba, Bart Bogaerts, and Tias Guns.
\newblock Efficiently explaining {CSP}s with unsatisfiable subset optimization.
\newblock {\em J. Artif. Intell. Res.}, 78:709--746, 2023.

\bibitem[GMN22]{GMN22AuditableConstraintProgrammingSolver}
Stephan Gocht, Ciaran McCreesh, and Jakob Nordstr{\"{o}}m.
\newblock An auditable constraint programming solver.
\newblock In {\em {CP}}, volume 235 of {\em LIPIcs}, pages 25:1--25:18. Schloss
  Dagstuhl, 2022.

\bibitem[GN21]{GN21CertifyingParityReasoningEfficientlyUsingPseudo-Boolean}
Stephan Gocht and Jakob Nordstr{\"{o}}m.
\newblock Certifying parity reasoning efficiently using pseudo-{Boolean}
  proofs.
\newblock In {\em {AAAI}}, pages 3768--3777. {AAAI} Press, 2021.

\bibitem[HSJ17]{ijcai/HyttinenSJ17}
Antti Hyttinen, Paul Saikko, and Matti J{\"{a}}rvisalo.
\newblock A core-guided approach to learning optimal causal graphs.
\newblock In {\em {IJCAI}}, pages 645--651. ijcai.org, 2017.

\bibitem[IMM16]{IMM16PropositionalAbductionImplicitHittingSets}
Alexey Ignatiev, Ant{\'{o}}nio Morgado, and Jo{\~{a}}o Marques{-}Silva.
\newblock Propositional abduction with implicit hitting sets.
\newblock In {\em {ECAI}}, volume 285 of {\em {FAIA}}, pages 1327--1335. {IOS}
  Press, 2016.

\bibitem[IPLM15]{IPLM15SmallestMUSExtractionMinimalHittingSet}
Alexey Ignatiev, Alessandro Previti, Mark~H. Liffiton, and Jo{\~{a}}o
  Marques{-}Silva.
\newblock Smallest {MUS} extraction with minimal hitting set dualization.
\newblock In {\em {CP}}, volume 9255 of {\em {LNCS}}, pages 173--182. Springer,
  2015.

\bibitem[IVS{\etalchar{+}}25]{IVSBBJ25Zenodo}
Hannes Ihalainen, Dieter Vandesande, André Schidler, Jeremias Berg, Bart
  Bogaerts, and Matti Järvisalo.
\newblock Experimental repository for ``efficient and reliable hitting-set
  computations for the implicit hitting set approach''.
\newblock \url{https://doi.org/10.5281/zenodo.17600095}, 2025.

\bibitem[IVS{\etalchar{+}}26]{IVSBBJ26EfficientReliableHitting-SetComputationsImplicitHitting}
Hannes Ihalainen, Dieter Vandesande, André Schidler, Jeremias Berg, Bart
  Bogaerts, and Matti Järvisalo.
\newblock Efficient and reliable hitting-set computations for the implicit
  hitting set approach.
\newblock In {\em Proceedings of The 40th Annual {AAAI} Conference on
  Artificial Intelligence}, 2026.
\newblock accepted for publication.

\bibitem[JBBJ25]{JBBJ25CertifyingParetoOptimalityMulti-ObjectiveMaximumSatisfiability}
Christoph Jabs, Jeremias Berg, Bart Bogaerts, and Matti J{\"{a}}rvisalo.
\newblock Certifying pareto optimality in multi-objective maximum
  satisfiability.
\newblock In {\em {TACAS} {(2)}}, volume 15697 of {\em {LNCS}}, pages 108--129.
  Springer, 2025.

\bibitem[LS08]{LS08AlgorithmsComputingMinimalUnsatisfiableSubsetsConstraints}
Mark~H. Liffiton and Karem~A. Sakallah.
\newblock Algorithms for computing minimal unsatisfiable subsets of
  constraints.
\newblock {\em J. Autom. Reason.}, 40(1):1--33, 2008.

\bibitem[NMBJ22]{NMBJ22ComputingSmallestMUSesQuantifiedBooleanFormulas}
Andreas Niskanen, Jere Mustonen, Jeremias Berg, and Matti J{\"{a}}rvisalo.
\newblock Computing smallest {MUS}es of quantified {Boolean} formulas.
\newblock In {\em {LPNMR}}, volume 13416 of {\em {LNCS}}, pages 301--314.
  Springer, 2022.

\bibitem[Oer25]{PBOxide}
Andy Oertel.
\newblock {PBOxide}: {V}erifier for pseudo-{B}oolean proofs rewritten in rust.
\newblock \url{https://gitlab.com/MIAOresearch/software/pboxide}, 2025.

\bibitem[RM09]{RM09Pseudo-BooleanCardinalityConstraints}
Olivier Roussel and Vasco~M. Manquinho.
\newblock Pseudo-{Boolean} and cardinality constraints.
\newblock In {\em Handbook of Satisfiability}, volume 185 of {\em {FAIA}},
  pages 695--733. {IOS} Press, 2009.

\bibitem[Rou24]{url:pbcomp2024}
Olivier Roussel.
\newblock {Pseudo-Boolean} competition 2024.
\newblock \url{https://www.cril.univ-artois.fr/PB24/}, 2024.

\bibitem[SBJ16]{SBJ16LMHSSAT-IPHybridMaxSATSolver}
Paul Saikko, Jeremias Berg, and Matti J{\"{a}}rvisalo.
\newblock {LMHS:} {A} {{SAT}-IP} hybrid {MaxSAT} solver.
\newblock In {\em {SAT}}, volume 9710 of {\em {LNCS}}, pages 539--546.
  Springer, 2016.

\bibitem[SBJ21]{SBJ21Pseudo-BooleanOptimizationImplicitHittingSets}
Pavel Smirnov, Jeremias Berg, and Matti J{\"{a}}rvisalo.
\newblock Pseudo-{Boolean} optimization by implicit hitting sets.
\newblock In {\em {CP}}, volume 210 of {\em LIPIcs}, pages 51:1--51:20. Schloss
  Dagstuhl, 2021.

\bibitem[SBJ22]{SBJ22ImprovementsImplicitHittingSetApproachPseudo-Boolean}
Pavel Smirnov, Jeremias Berg, and Matti J{\"{a}}rvisalo.
\newblock Improvements to the implicit hitting set approach to pseudo-{Boolean}
  optimization.
\newblock In {\em {SAT}}, volume 236 of {\em LIPIcs}, pages 13:1--13:18.
  Schloss Dagstuhl, 2022.

\bibitem[SDAJ18]{SDAJ18HybridApproachOptimizationAnswerSetProgramming}
Paul Saikko, Carmine Dodaro, Mario Alviano, and Matti J{\"{a}}rvisalo.
\newblock A hybrid approach to optimization in answer set programming.
\newblock In {\em {KR}}, pages 32--41. {AAAI} Press, 2018.

\bibitem[Smi21]{SmirnovMSc}
Pavel Smirnov.
\newblock Pseudo-boolean optimization by implicit hitting sets.
\newblock Master's thesis, University of Helsinki, 2021.

\bibitem[SWJ16]{SWJ16ImplicitHittingSetAlgorithmsReasoningBeyond}
Paul Saikko, Johannes~Peter Wallner, and Matti J{\"{a}}rvisalo.
\newblock Implicit hitting set algorithms for reasoning beyond {NP}.
\newblock In {\em {KR}}, pages 104--113. {AAAI} Press, 2016.

\bibitem[VCB26]{VCB26CertifiedBranch-and-BoundMaxSATSolving}
Dieter Vandesande, Jordi Coll, and Bart Bogaerts.
\newblock Certified branch-and-bound {MaxSAT} solving.
\newblock In {\em Proceedings of The 40th Annual {AAAI} Conference on
  Artificial Intelligence}, 2026.
\newblock accepted for publication.

\bibitem[VDB22]{VDB22QMaxSATpbCertifiedMaxSATSolver}
Dieter Vandesande, Wolf {De Wulf}, and Bart Bogaerts.
\newblock {QMaxSATpb}: {A} certified {MaxSAT} solver.
\newblock In {\em {LPNMR}}, volume 13416 of {\em {LNCS}}, pages 429--442.
  Springer, 2022.

\end{thebibliography}
\end{document}